\documentclass{article}
\usepackage{array}
\usepackage{graphicx}
\usepackage{amsfonts}
\usepackage{xcolor}
\usepackage{wrapfig}
\usepackage{booktabs}
\usepackage{subcaption}

\usepackage[toc,page,header]{appendix}
\usepackage{minitoc}
\usepackage{color, soul} 

\makeatletter
\newcommand{\appendixcontents}{
    \clearpage
    \section*{Appendix Contents}
    \@starttoc{atoc}
}
\makeatother

\usepackage[table]{xcolor}

\newcommand{\web}{https://yingyuan0414.github.io/grasp2dexterity/}
\usepackage[final]{corl_2026} 

\title{From Grasps to Dexterity: Large-Scale Grasp Pretraining for Dexterous Manipulation}

\author{
  Ying Yuan\textsuperscript{*}, Xinyu Liu\textsuperscript{*}, Sriram Krishna, David Held\\
  Robotics Institute, Carnegie Mellon University\\
  \textsuperscript{*}Equal contribution
}

\begin{document}
\doparttoc 
\faketableofcontents
\maketitle


\vspace{-2.8em}

\vspace{-1.0em}
\begin{figure}[!h]
    \centering
    \includegraphics[width=\linewidth]{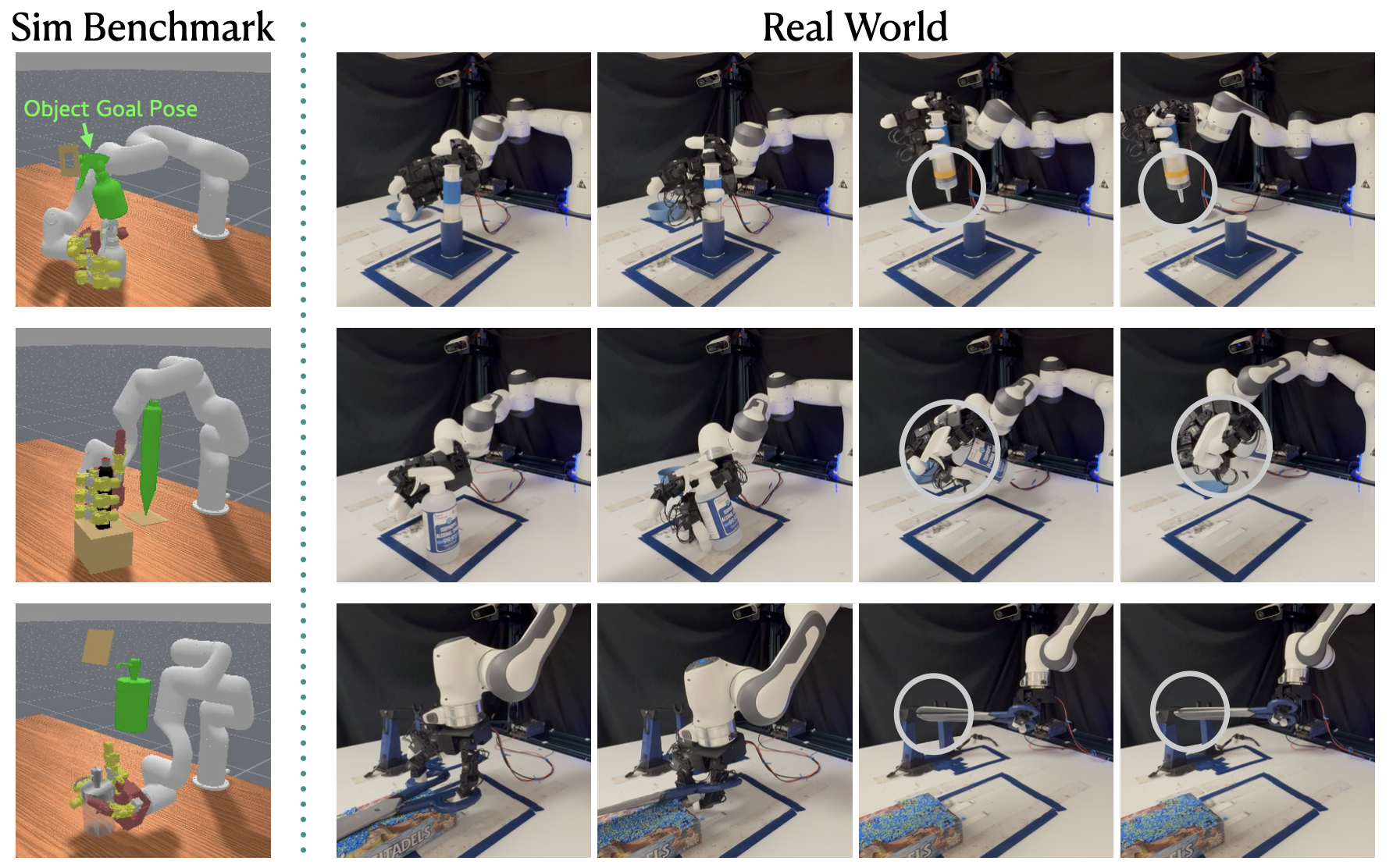}
    \caption{\small{\textbf{Left}: Our simulation benchmark, \texttt{DexCraft}, with articulated tool use tasks. We visualize object goal poses with green object meshes. \textbf{Right}: With a real-world robot, our policy can perform highly dexterous tasks using proprioception and RGB-D perception as feedback. More videos are available on our \href{\web}{project website}.}}
    \label{fig:teaser}
    \vspace{-1.5em}
\end{figure}

\begin{abstract}
    Large-scale dexterous grasp datasets encode rich priors over hand-object interaction, but their use has largely been confined to grasp generation and pick-and-place manipulation. We study whether such data can instead support functional dexterity in articulated tool use, where a robot must acquire a tool, maintain contact, and operate its functional moving parts. We adapt a hierarchical imitation learning framework that combines high-level hand sub-goal prediction with a low-level goal-conditioned controller. We construct a 355k-trajectory grasp-pretraining dataset from large-scale dexterous grasp annotations and use it to pretrain the low-level controller. The controller is then fine-tuned on downstream task demonstrations. To evaluate this setting, we introduce DexCraft, a simulation benchmark with six articulated tool-use tasks requiring coordinated finger motion. Across simulation and real-world experiments, our approach outperforms end-to-end diffusion policy baselines and hierarchical policies trained from scratch. In the real world, it improves full-task success by 33.3 percentage points over DP3. These results show that grasp datasets can serve not only as resources for grasp synthesis, but also as scalable pretraining data for contact-rich dexterous manipulation. Videos are shown on our \href{\web}{project website}.
\end{abstract}




\section{Introduction}

Dexterous manipulation is essential for robot autonomy, enabling functional tasks that require coordinated finger motion, stable contact, and precise force application.
Large-scale robot grasping datasets have recently emerged as a promising resource for learning dexterous manipulation. Recent grasping datasets~\cite{wang2022dexgraspnet, chen2025dexonomy, zhang2024graspxl, ye2025dex1b} generate millions of physically plausible grasps over diverse object categories. These datasets have been widely used to train grasp prediction models~\cite{wang2022dexgraspnet, chen2025dexonomy, zhang2024dexgraspnet, 10753039} and closed-loop grasping policies~\cite{ye2025dex1b, xu2023unidexgrasp, wan2023unidexgrasp++}.
However, their use has largely been limited to grasp generation and pick-and-place settings. Much less attention has been paid to whether grasp datasets can support longer-horizon downstream tasks, where the robot must not only grasp an object, such as a syringe, but also use it to accomplish a functional goal.

Functional tool use poses challenges that go beyond conventional grasping. Many tool use tasks require multiple stages of interaction: acquiring the tool, maintaining a stable grasp during contact-rich motion, and adapting the grasp as the task progresses. Articulated tool use makes the problem even harder, since the robot must both preserve the relevant object pose and apply force at the right contact location to move the articulated part. In this paper, we investigate a key question: \textit{How can we leverage large-scale grasp datasets to achieve robot dexterity in downstream tasks like articulated tool use?} 

To study this problem systematically, we introduce \texttt{DexCraft}, a simulation benchmark for dexterous articulated tool use. While prior benchmarks~\cite{Rajeswaran-RSS-18,bao2023dexart} have advanced research on learning to manipulate articulated objects, they do not explicitly require the fine-grained, coordinated hand motions needed for highly dexterous tool operation.
\texttt{DexCraft} fills this gap with six tool use tasks designed to require high levels of dexterity beyond the capabilities of parallel-jaw grippers, e.g. holding a spray bottle and squeezing its trigger. 

Towards pretraining a low-level controller to solve these tasks, we also construct \texttt{G2D-Pretrain}, a large-scale dataset of 355k grasp trajectories by augmenting grasp annotations from Dexonomy~\cite{chen2025dexonomy}. These trajectories expose the controller to diverse hand-object contact patterns, grasp formations, and post-grasp motions before it is fine-tuned on downstream tool-use demonstrations.



However, grasp trajectories alone do not specify which functional outcome the robot should achieve, how the task should progress, or when the hand should transition from grasping to tool operation. Hierarchical policy learning has shown promise in manipulation with parallel-jaw grippers~\cite{wang2025articubot,krishna2026ghost}. We therefore combine grasp-pretrained low-level control with hierarchical policy learning, where a high-level policy predicts task-relevant hand sub-goals and the low-level controller executes them.


In summary, our contributions are as follows:
1) We introduce \texttt{DexCraft}, a simulation benchmark that consists of six articulated tool use tasks for dexterous manipulation.
2) We construct \texttt{G2D-Pretrain}, a large-scale hand-object interaction demonstration dataset augmented from Dexonomy~\cite{chen2025dexonomy} containing 355k demonstration trajectories.
3) We propose a pipeline that leverages a large-scale grasp synthesis dataset and a hierarchical policy representation, enabling data-efficient learning of dexterous tasks.
4) We show that our method improves the average success rate by 33.3 percentage points over the DP3 baseline on real-world articulated tool use tasks.




    

\section{Related Work}

\textbf{Dexterous Grasping and Tool Use Manipulation.} Dexterous grasping aims to generate stable hand-object contacts for multi-fingered robot hands. Classical methods synthesize grasps through optimization~\cite{Ciocarlie2007DexterousGV,graspit,grasp_synthesis_clutter}, while later learning-based methods use contact maps, object-centric affordances, or learned grasp representations to synthesize dexterous grasps~\cite{contactopt,Mandikal2020LearningDG,contactgrasp,gendexgrasp}. 
Recent works further scale dexterous grasp synthesis with differentiable simulation and large datasets~\cite{graspd,wang2022dexgraspnet,chen2025dexonomy,zhang2024graspxl,ye2025dex1b,zhang2024dexgraspnet,10753039}. However, these methods do not directly address how grasps should support functional manipulation.
Dexterous tool use requires reasoning about task progress and changing contacts. Prior methods learn from robot demonstrations~\cite{Seita2022toolflownet,qitooluse2024,lin2024learning} but such data is costly. Others infer affordances from human demonstrations or videos~\cite{10.1007/978-3-030-95459-8_9,agarwal2023dexterous,Hadjivelichkov2022affcorrs,bahl2023affordances,affordance_diffusion}, often requiring retargeting or separate controllers. 
RL-based methods~\cite{Rajeswaran-RSS-18,dexpoint,chen2023visual,qi2023general,wang2024penspin} can learn complex skills, but require task-specific reward design. 
DexGen~\cite{inproceedings}  provides a complementary direction, pretraining a dexterous controller to refine coarse commands from teleoperation. In contrast, our work learns an autonomous visuomotor policy for articulated tool use, where the robot predicts task-relevant hand sub-goals from perception and executes them with a low-level controller.



\textbf{Hierarchical Policy Learning.} 
Hierarchical reinforcement learning studies temporal abstraction for long-horizon decision-making~\cite{sutton1999between,dietterich2000hierarchical,pmlr-v70-vezhnevets17a,nachum2018data}. 
Our work instead focuses on hierarchical imitation learning, where hierarchy structures policies learned from demonstrations.
MimicGen~\cite{mandlekar2023mimicgen} uses hierarchy in data generation, while Amplify~\cite{collins2025amplify} introduces an intermediate abstraction by separating visual motion prediction from action inference.
Closest to our setting, recent hierarchical imitation policies~\cite{wang2025articubot,krishna2026ghost} build on diffusion-based architectures~\cite{chi2023diffusionpolicy,Ze2024DP3} and decompose control into a high-level sub-goal prediction and a low-level controller. However, they use sub-goal representations designed for parallel-jaw grippers, whereas dexterous hands require richer representations.




\textbf{Dexterous Manipulation Datasets and Benchmarks.} Existing benchmarks such as Adroit~\cite{Rajeswaran-RSS-18} and DexArt~\cite{bao2023dexart} focus on in-hand or articulated object manipulation. In contrast, \texttt{DexCraft} focuses on contact-rich tasks of articulated tool use, complementing prior benchmarks.
Large-scale dexterous grasp datasets~\cite{wang2022dexgraspnet,ye2025dex1b,zhang2024dexgraspnet,he2025dexvlg} provide useful priors for hand-object interaction. GraspXL~\cite{zhang2024graspxl} generates grasping motions with various graspable regions and approach directions, while Dexonomy~\cite{chen2025dexonomy} synthesizes diverse taxonomy-labeled grasps with diverse contact patterns and hand configurations.
Our \texttt{G2D-Pretrain} dataset builds on Dexonomy~\cite{chen2025dexonomy} by converting diverse, physically plausible grasp annotations into demonstration trajectories for low-level policy pretraining, bridging grasp synthesis data and downstream dexterous tool-use behavior.

\section{Problem Statement and Assumptions}
\label{section:assumption}
We consider dexterous manipulation tasks as a Markov Decision Process (MDP), and learn a robot manipulation policy $\pi$ that maps the state space $\mathcal{S}$ to the action space $\mathcal{A}$.  We assume access to a dataset with $N$ demonstrations $\mathcal{D}=\{s_0^i, a_0^i, s_1^i, a_1^i, \cdots, s_{T_i}^i\}_{i=1}^N$, where each $s_0^i$ is sampled from the initial state distribution. 
We also assume the dexterous manipulation tasks can be decomposed into sequential phases. Each task consists of sub-goal frames $g_1, g_2, \cdots, g_m\in \{0, \cdots, T_i\}$. These frames correspond to important transitions in the demonstration, for example, when the robot changes its grasp configuration or when the end effector moves to a new target pose. 

\section{DexCraft Benchmark}
\label{section:benchmark}

To systematically study the problem of articulated tool use, we propose the \texttt{DexCraft} benchmark which contains six dexterous manipulation tasks involving articulated tool use. In each task, the robot hand needs to grasp the articulated tool, lift it to a target pose, and perform the function of the tool. We place a reference object that the tool will interact with at a fixed offset from the tool's goal pose.
An episode is considered successful when both: 1) the position and the orientation of the tool match the tool's goal pose, and 2) the 
articulated joint of the object is triggered beyond a threshold.

\begin{wrapfigure}{r}{0.6\linewidth}
    \centering
    \vspace{-0.8em}
    \includegraphics[width=\linewidth]{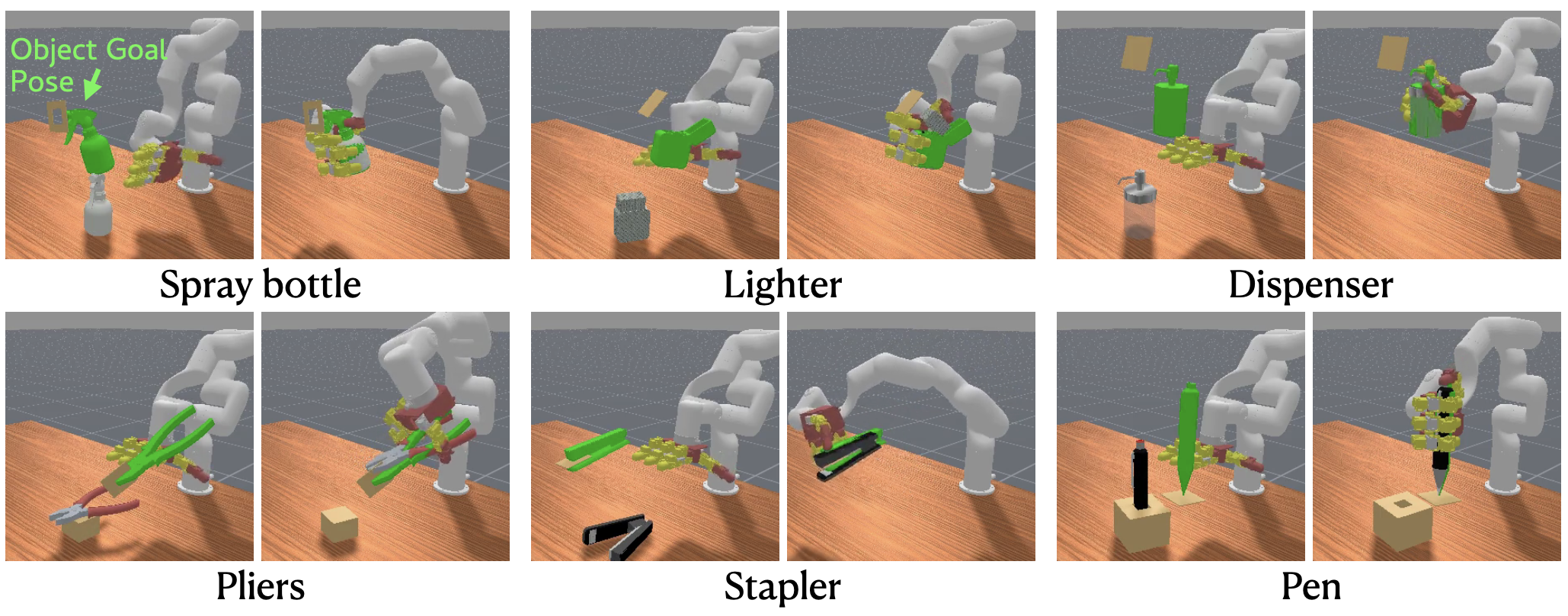}
    \caption{\small{Visualization of \texttt{DexCraft} tasks. For each task, the initial frame is shown on the left and the target frame is on the right. We visualize target object positions with green object meshes (unobserved by the policy). Reference objects that the tools will interact with are placed relative to the goal. The robot hand is required to grasp the object, lift it to the target pose, and trigger the object's articulated joint.}}
    \label{fig:benchmark}
    \vspace{-1.0em}
\end{wrapfigure}

Visualizations of the tasks are shown in Figure~\ref{fig:benchmark}. 
The tasks are defined as follows:
1) \textbf{\textit{Spray bottle}}: The robot grasps a spray bottle, moves it to the goal pose in front of a cloth, and squeezes the trigger with its index finger.
2) \textbf{\textit{Lighter}}: The robot grasps a flip-top lighter, moves it to the goal pose indicated by a reference object, and flips open the lid with its thumb.
3) \textbf{\textit{Dispenser}}: The robot grasps a pump dispenser, moves it to the goal pose indicated by a reference object, and presses the pump with its index finger.
4) \textbf{\textit{Pliers}}: The robot grasps a pair of pliers, moves them to the goal pose with a reference object positioned between the jaws, and closes the jaws.
5) \textbf{\textit{Stapler}}: The robot grasps a stapler, moves it to the goal pose with a reference object positioned between the top arm and the base, and presses the top arm.
6) \textbf{\textit{Pen}}: The robot grasps a pen, moves it to the goal pose indicated by a reference object near the pen tip, and clicks the button with its thumb.


\textbf{Environment Setup.} We implement our tasks in the ManiSkill3 physical simulator~\cite{taomaniskill3} using an XArm6 robot arm (6 DoF) with a LEAP Hand~\cite{shaw2023leaphand} (16 DoF). We use articulated assets from the PartNet-Mobility Dataset~\cite{Xiang_2020_SAPIEN,Mo_2019_CVPR,chang2015shapenet}. \texttt{DexCraft} focuses on functional dexterity under object pose variation rather than category-level generalization. We thus use one asset for each task and randomly sample the object's initial pose and target pose. The observation space includes the robot's joint position $q_{pos}\in \mathbb{R}^{22}$, joint velocity $q_{vel}\in \mathbb{R}^{22}$, previous joint position target $\bar{q}_{pos}\in \mathbb{R}^{22}$, end-effector's pose $p_{eef}\in \mathbb{R}^7$, the observed point cloud $P_o\in\mathbb{R}^{1024\times 3}$ and the hand skeleton points $P_{hand}\in \mathbb{R}^{16\times 3}$. The action space is a 22-dimensional vector that consists of 6-DoF for the arm and 16-DoF for the hand. We use delta end-effector control for the arm and delta joint position control for the hand joints. 

\textbf{Data Collection.} We use keyboard teleoperation to collect 5 demos for each task. For end-effector control, we use 6 keys to teleoperate the translation of the end-effector, and another 6 keys to rotate it by roll, yaw, and pitch angles. For hand control, we design finger primitives to achieve different grasps. For example, we map two keys to bending and straightening each finger. We also map certain keys to controlling each joint on the thumb so that the hand can achieve various grasps. We then use MimicGen~\cite{mandlekar2023mimicgen} to augment manually-collected demos to 100 demos with more spatial variations.

\section{From Large-Scale Grasp Data to Dexterity}

\begin{figure}[t]
    \centering
    \includegraphics[width=\linewidth]{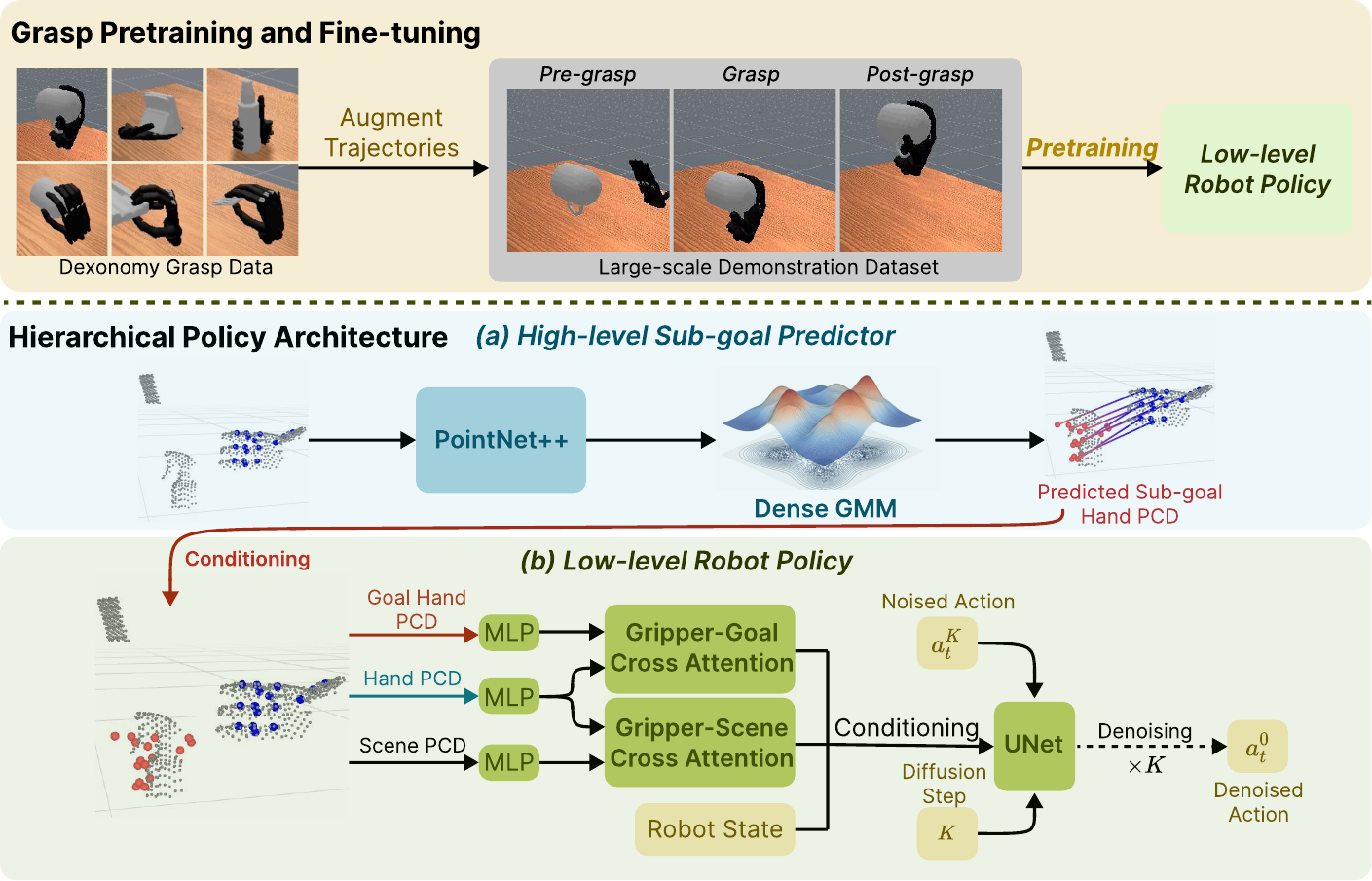}
    \caption{\small{Our method integrates large-scale grasp pretraining with a hierarchical policy framework. (a) A \textbf{high-level sub-goal prediction policy} takes the current point cloud observation as input and predicts the positions of hand key points. (b) A \textbf{low-level policy} is conditioned on predicted sub-goal key points and current observation and predicts action chunks for the controller. \textbf{Top:} We augment the Dexonomy~\cite{chen2025dexonomy} dataset into grasp trajectories and pretrain the low-level policy, which we fine-tune with data from the downstream task.}}
    \label{fig:pipeline}
    \vspace{-1em}
\end{figure}

\subsection{Policy Learning with Hierarchical Policy Representation}
\label{section:5.1}
Previous works~\cite{wang2025articubot} use a hierarchical policy representation and show sample efficiency in parallel-jaw gripper tasks. In this section, we introduce how we adapt the hierarchical policy framework and extend it to general dexterous manipulation. 

\textbf{High-Level Sub-Goal Prediction Policy.} The high-level policy $\pi_{high}$ predicts an explicit 3D representation of the end-effector state $e_{g_k}$ at the $k$-th sub-goal time step $g_k$, given the current point cloud observation. In previous works~\cite{wang2025articubot,krishna2026ghost}, the sub-goal is defined as the positions of 4 points on the parallel-jaw gripper. However, for high-DoF robot hands, four points are insufficient to describe the full hand state: the same palm and fingertip poses can correspond to different internal joint configurations, contact modes, and functional finger roles. We therefore predict 16 hand keypoints for LEAP-hand manipulation, with one keypoint on each hand link, as shown in Figure~\ref{fig:pipeline}.

Following prior work~\cite{wang2025articubot}, we use PointNet++~\cite{qi2017pointnetplusplus} as the backbone and model the goal distributions as a dense Gaussian Mixture Model (GMM). We train $\pi_{high}$ by minimizing the loss $\mathcal{L}_{high}=\mathbb{E}_{(o_t,e_{g(t)})\sim \mathcal{D}}||\pi_{high}(o_t)-e_{g(t)}||_2^2$, where $e_{g(t)}\in \mathbb{R}^{16\times3}$ is the sub-goal hand keypoints at time step $t$. The point cloud observation $o_t$ consists of: 1) point cloud $o_{cam}\in \mathbb{R}^{N_{cam}\times 3}$ from the depth camera, 2) current hand keypoints $e_t\in \mathbb{R}^{16\times 3}$. We append one-hot vectors as labels to each point and concatenate the point clouds together as input.

\textbf{Low-Level Goal Conditioned Policy.} The low-level policy $\pi_{low}$ predicts the delta transformation of the end-effector and delta joint control of the fingers, given the point cloud observation $o_t$, the low-dimensional robot state $s_t$, and the sub-goal hand keypoints $e_{g(t)}$. We adopt a modified 3D Diffusion Policy (DP3) architecture as in~\cite{wang2025articubot}, which includes gripper-goal cross-attention and gripper-scene cross-attention as shown in Figure~\ref{fig:pipeline}. 
We train $\pi_{low}$ to minimize the behavior cloning loss $\mathcal{L}=\mathbb{E}_{(o_t,s_t,e_{g(t)},a_t)\sim \mathcal{D}}||\pi_{low}(o_t, s_t, e_{g(t)})-a_t||_2^2$, where $a_t$ is the delta transformation of the end-effector, including the translation, rotation, and hand joint movement.

We empirically divide demonstrations into three stages: grasping, aligning, and actuating. The sub-goal frames consist of the last frame of each stage.
We train $\pi_{high}$ and $\pi_{low}$ independently, using ground-truth sub-goals from demonstrations when training both. At inference time, we execute $\pi_{high}$ and $\pi_{low}$ sequentially at the same frequency, i.e. $\pi_{high}$ predicts a sub-goal conditioned on the current observation and $\pi_{low}$ predicts an action chunk conditioned on the sub-goal prediction.

\subsection{Augmenting Dexonomy Dataset for Low-Level Policy Pretraining}


The Dexonomy dataset~\cite{chen2025dexonomy} contains 10.7k objects and 9.5M grasps of a Shadow hand~\cite{Sharma_Tokas_Puri_Sharda_2014}, which covers 31 grasp types in the GRASP taxonomy~\cite{grasp_taxonomy}. This diversity of grasp types is helpful for dexterous manipulation of articulated tool use in which various grasp types might be needed throughout the manipulation tasks.  Further, the 22-DoF Shadow hand used in this dataset is one of the most kinematically complex robot hands to date and has been shown to transfer to lower-DoF hands with reliable performance~\cite{wei2024dro,wei2026handruleallcanonical}. 
In this section, we introduce how we augment the large-scale grasp dataset into a demonstration dataset, \texttt{G2D-Pretrain}, and use it to pretrain our low-level policy. 


\textbf{Augmenting Dexonomy into Demonstration Dataset.} Each data point in Dexonomy~\cite{chen2025dexonomy} includes three key poses of the hand, i.e. pre-grasp pose, grasp pose, and squeeze pose. However, to pretrain the low-level policy, we need complete trajectories.  
Thus, as shown in the top part of Figure~\ref{fig:pipeline}, we augment the Dexonomy dataset into 3-stage trajectories: 1) \textit{Pre-grasp}: Starting from a random pose, the robot hand moves to approach the pre-grasp pose with motion planning. 2) \textit{Grasp}: Starting from the pre-grasp pose, the robot hand is controlled to reach the grasp pose and then the squeeze pose with interpolation. 3) \textit{Post-grasp}: After squeezing the object, the robot hand translates to another random position in the scene with interpolation. We use the ManiSkill3 simulator~\cite{taomaniskill3} with a floating Shadow hand to physically validate the generated trajectories. We verify that the robot hand is grasping the object in the last frame of the simulation. We filter out generated trajectories that do not meet this criteria, and we thereby obtain 355k successful trajectories for use for pretraining. 

\textbf{Grasp Pretraining and Downstream Fine-tuning for Low-Level Policy.} We pretrain our low-level policy $\pi_{low}$, as described in Section~\ref{section:5.1}, with the large-scale grasp demonstration dataset \texttt{G2D-Pretrain}. The low-dimensional observation space and the action space in \texttt{G2D-Pretrain} can be different from those in downstream tasks due to differences in robot morphology. 
Specifically, our observation space includes the robot's joint position $q_{pos}$, joint velocity $q_{vel}$, previous joint position target $\bar{q}_{pos}$ and the end-effector pose $p_{eef}$. 
If a downstream task is missing an entry or the state observation shape does not match the pretrained model, we zero-pad the states at the corresponding entries when fine-tuning. The  action space consists of a 6-DoF delta end-effector pose and the delta joint angles for a 22-DoF hand following the joint order of the Shadow hand. For downstream tasks with different hands such as LEAP hand, we map the joint entries semantically across robot morphologies and zero-pad the unused entries.

\section{Experiments}
\label{sec:result}

In this section, we compare our approach for articulated tool use to several baselines in both simulation and the real world. Specifically, we study 1) whether hierarchical policies enable data-efficient learning for dexterous manipulation tasks, 2) the effect of low-level policy pretraining on the overall performance, and 3) critical design choices in the system.

\subsection{Experiment Setup}
\textbf{Task Description.} For simulation evaluation, we use our \texttt{DexCraft} benchmark as described in Section~\ref{section:benchmark}. For real world experiments, we evaluate our approach on three manipulation tasks: \textit{syringe}, \textit{spray bottle}, and \textit{scissors}. 
For the \textit{syringe} task, 
the robot needs to grasp the syringe, move it to the blue bowl, and push the plunger to the 0 mL mark. For the \textit{spray bottle} task, the robot needs to grasp the spray bottle, move it to the blue bowl and use the index finger to squeeze the trigger. For the \textit{scissors} task, the robot needs to precisely grasp the toy scissors and move it to the cable.
We randomly reset the object inside the blue area during data collection and policy evaluation.

\begin{wrapfigure}{r}{0.5\linewidth}
    \centering
    \vspace{-0.8em}
    \includegraphics[width=\linewidth]{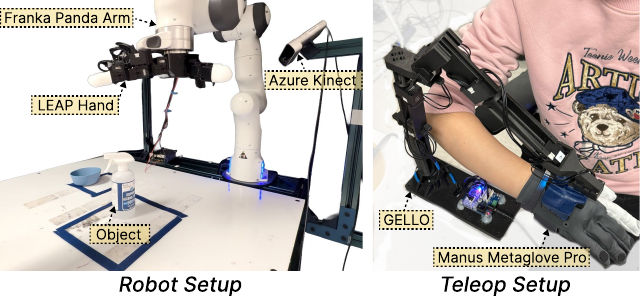}
    \caption{Real World Setup}
    \label{fig:real-setup}
    \vspace{-1.0em}
\end{wrapfigure}

\textbf{Environment Setup and Data Collection.} The simulation setup is detailed in Section~\ref{section:benchmark}. For real world tasks, shown in Figure~\ref{fig:real-setup}, we use a Franka Emika Panda Robot with a LEAP Hand, and two Microsoft Azure Kinect cameras for visual perception. For data collection, we attach a Manus Metaglove Pro to GELLO~\cite{wu2023gello} and teleoperate the system at 30Hz. We collect 50 demos for each task, manually annotate sub-goal frames, and then train and deploy our policies at 15Hz.

\textbf{Evaluation Metrics.} 
In simulation, we evaluate full task success rate over evaluation episodes. 
For real-world experiments, we report both full-task success and stage-wise success to better capture partial progress in long-horizon manipulation. 
Each task is decomposed into three sequential milestones: grasping the tool, aligning it with the target, and triggering the articulated component. 
A later stage is counted as successful only if all preceding stages are completed. 
This metric helps distinguish failures in object acquisition, tool positioning, and final actuation.

\subsection{Baselines and Ablations}
We compare our method to the following baselines: 
(1) \textit{Diffusion Policy (DP)}~\cite{chi2023diffusionpolicy}, a visuomotor policy that takes RGB images as input and generates robot action sequences through a denoising diffusion process.
(2) \textit{3D Diffusion Policy (DP3)}~\cite{Ze2024DP3}, a diffusion policy that takes 3D point cloud as input and outputs delta end-effector transformations as the actions. 
(3) \textit{UniDex-VLA}~\cite{zhang2026unidex}, a recent 3D VLA policy pretrained on UniDex-Dataset.
(4) \textit{Dex1B}~\cite{ye2025dex1b}, a lightweight CVAE-based generative model trained from scratch.

We also experiment on the following ablations to study critical design choices in the system: 
(1) \textit{Pretrained DP3 (DP3-PT)}, pretraining the end-to-end DP3 policy with \texttt{G2D-Pretrain} and finetuning with downstream task data.
(2) \textit{Hierarchical Policy Learning from Scratch (HP-Scratch)}, essentially our method without grasp pretraining. 
(3) \textit{Pretrain w/ GraspXL (GraspXL-PT)}: Using GraspXL~\cite{zhang2024graspxl} with Allegro Hand motions instead of using \texttt{G2D-Pretrain} to pretrain the low-level policy.
(4) \textit{PT-215k}, pretraining with a subset of 215k trajectories of our \texttt{G2D-Pretrain} dataset.
(5) \textit{PT w/o Approach}, pretraining with only grasping and lifting stages of synthesized trajectories in our \texttt{G2D-Pretrain} dataset.
(6) \textit{PT w/o Lift}, pretraining with only approaching and grasping stages of synthesized trajectories in our \texttt{G2D-Pretrain} dataset.

\subsection{Simulation Results}

\begin{table*}[t]
\centering
\scriptsize
\setlength{\tabcolsep}{2.3pt}
\begin{tabular}{
lccc|ccc|ccc|ccc|ccc|ccc||
>{\columncolor{gray!15}}c
>{\columncolor{gray!15}}c
>{\columncolor{gray!15}}c
}
\toprule
& \multicolumn{3}{c|}{SprayBot.}
& \multicolumn{3}{c|}{Lighter}
& \multicolumn{3}{c|}{Dispenser}
& \multicolumn{3}{c|}{Pliers}
& \multicolumn{3}{c|}{Stapler}
& \multicolumn{3}{c||}{Pen}
& \multicolumn{3}{>{\columncolor{gray!10}}c}{\textbf{Avg.}} \\
\cmidrule(lr){2-4}
\cmidrule(lr){5-7}
\cmidrule(lr){8-10}
\cmidrule(lr){11-13}
\cmidrule(lr){14-16}
\cmidrule(lr){17-19}
\cmidrule(l){20-22}
\# demo
& 20 & 50 & 100
& 20 & 50 & 100
& 20 & 50 & 100
& 20 & 50 & 100
& 20 & 50 & 100
& 20 & 50 & 100
& 20 & 50 & 100 \\
\midrule
DP
& 0.4 & 0.4 & 0.8
& 2.0 & 2.4 & 3.8
& 0.6 & 0.8 & 1.8
& 0.4 & 1.8 & 2.2
& 0.0 & 1.6 & 2.2
& 2.8 & 3.6 & 3.2
& 1.0 & 1.8 & 2.3 \\

DP3
& 2.4 & 4.8 & 21.0
& 3.2 & 8.0 & 15.8
& 1.8 & 4.4 & 14.2
& 2.4 & 7.6 & 13.4
& 2.4 & 20.0 & 39.2
& 2.2 & 5.6 & 10.4
& 2.4 & 8.4 & 19.1 \\

DP3-PT
& 7.6 & 15.2 & 29.0
& 11.2 & 26.6 & 35.2
& 11.4 & 5.8 & 17.4
& 10.4 & 21.2 & 23.2
& 10.4 & 38.8 & 43.8
& 9.2 & 25.8 & 29.6
& 10.1 & 22.3 & 29.7 \\
\midrule
GraspXL-PT 
& 16.8 & 44.0 & 44.2 
& 7.4 & \underline{34.6} & 36.8 
& 3.2 & 42.0 & 58.8 
& 15.8 & 32.6 & 38.8 
& 21.0 & 43.2 & \underline{71.6} 
& 13.6 & 31.4 & 42.0 
& 13.0 & 37.8 & 48.7 \\

HP-Scratch
& \underline{43.6} & \underline{58.6} & \underline{57.0}
& \underline{29.8} & 27.8 & \underline{44.2}
& \underline{43.8} & \underline{64.2} & \underline{73.8}
& \underline{31.6} & \underline{50.0} & \underline{54.2}
& \underline{34.2} & \underline{61.8} & 70.8
& \textbf{48.8} & \underline{31.0} & \underline{47.6}
& \underline{38.7} & \underline{48.9} & \underline{58.0} \\

Ours
& \textbf{59.2} & \textbf{61.4} & \textbf{68.4}
& \textbf{33.2} & \textbf{46.4} & \textbf{45.4}
& \textbf{54.0} & \textbf{72.8} & \textbf{78.6}
& \textbf{32.8} & \textbf{58.0} & \textbf{62.6}
& \textbf{49.2} & \textbf{66.6} & \textbf{77.2}
& \underline{43.8} & \textbf{56.4} & \textbf{54.6}
& \textbf{45.4} & \textbf{60.3} & \textbf{64.6} \\
\bottomrule
\end{tabular}
\caption{Success rate (\%) across six simulation tasks with 20, 50, and 100 demonstrations. The Avg. columns report the average over all tasks. We use 500 evaluation episodes in each setting.}
\label{tab:main_results}
\end{table*}

\begin{table}[t]
\centering
\scriptsize
\setlength{\tabcolsep}{2.8pt}
\renewcommand{\arraystretch}{1.05}

\begin{tabular}{lrrrrrrr}
\toprule
\textbf{Success (\%)}& \textbf{SprayBot.} & \textbf{Lighter} & \textbf{Dispenser} &
\textbf{Pliers} & \textbf{Stapler} & \textbf{Pen} &
\cellcolor{gray!15}\textbf{Avg.} \\
\midrule
Contact               & 98.8 & 98.2 & 100.0 & 99.8 & 96.6 & 99.4 & \cellcolor{gray!15}98.8 \\
Grasp                 & 58.2 & 32.0 & 55.8  & 71.6 & 61.0 & 95.0 & \cellcolor{gray!15}62.3 \\
Lift                  & 28.8 & 3.6  & 0.4   & 44.4 & 14.0 & 65.6 & \cellcolor{gray!15}26.1 \\
At Goal               & 7.6  & 0.6  & 0.0   & 4.4  & 0.0  & 12.2 & \cellcolor{gray!15}4.1 \\
Articulate after Lift & 6.4  & 0.0  & 0.0   & 2.8  & 0.0  & 9.2  & \cellcolor{gray!15}3.1 \\
\midrule
Full Success (UniDex-VLA~\cite{zhang2026unidex})
                      & 2.2 & 0.0 & 0.0 & 0.0 & 0.0 & 0.6 & \cellcolor{gray!15}0.5 \\
Full Success (Ours)   & 61.4 & 46.4 & 72.8 & 58.0 & 66.6 & 56.4 & \cellcolor{gray!15}60.3 \\
\bottomrule
\end{tabular}
\vspace{1em}
\caption{Comparison of our method to UniDex-VLA~\cite{zhang2026unidex}. Top: Stage-wise performance of UniDex-VLA
fine-tuned with 100 demos, 50 epochs per task in simulation. Bottom: Overall success rate of UniDex-VLA and our method with 100 demos, 50 epochs per task.}
\label{tab:object-stage-performance}
\vspace{-2.5em}
\end{table}

\begin{figure*}[t]
\centering
\begin{minipage}[t]{0.65\textwidth}
\vspace{5mm}
\centering
\scriptsize
\setlength{\tabcolsep}{2.5pt}
\renewcommand{\arraystretch}{1.05}

\begin{tabular}{lc|rrrrrrr}
\toprule
\textbf{Success (\%)}& \textbf{\# demo} &
\textbf{SprayBot.} & \textbf{Lighter} & \textbf{Dispenser} &
\textbf{Pliers} & \textbf{Stapler} & \textbf{Pen} &
\cellcolor{gray!15}\textbf{Avg.}\\
\midrule
Dex1B        & 100 & 1.9  & 3.6  & 0.0  & 6.1  & 2.8  & 2.9  & \cellcolor{gray!15}2.9 \\
PT-215k      & 50  & 56.7 & 33.2 & 65.5 & 52.4 & 66.4 & 59.4 & \cellcolor{gray!15}55.6 \\
PT w/o Appr. & 50  & 55.4 & 42.7 & 71.2 & 44.4 & 51.1 & 52.2 & \cellcolor{gray!15}52.8 \\
PT w/o Lift  & 50  & 68.6 & 40.3 & 67.3 & 55.6 & 56.5 & 53.3 & \cellcolor{gray!15}56.9 \\
\midrule
Ours         & 50  & 61.4 & 46.4 & 72.8 & 58.0 & 66.6 & 56.4 & \cellcolor{gray!15}60.3 \\
\bottomrule
\end{tabular}

\captionof{table}{Comparison of our method to Dex1B~\cite{ye2025dex1b} architecture and ablation on synthetic pretraining trajectories.}
\label{tab:more-ablations}
\end{minipage}
\hfill
\begin{minipage}[t]{0.32\textwidth}
\vspace{0.1mm}
\centering
\includegraphics[width=0.9\linewidth]{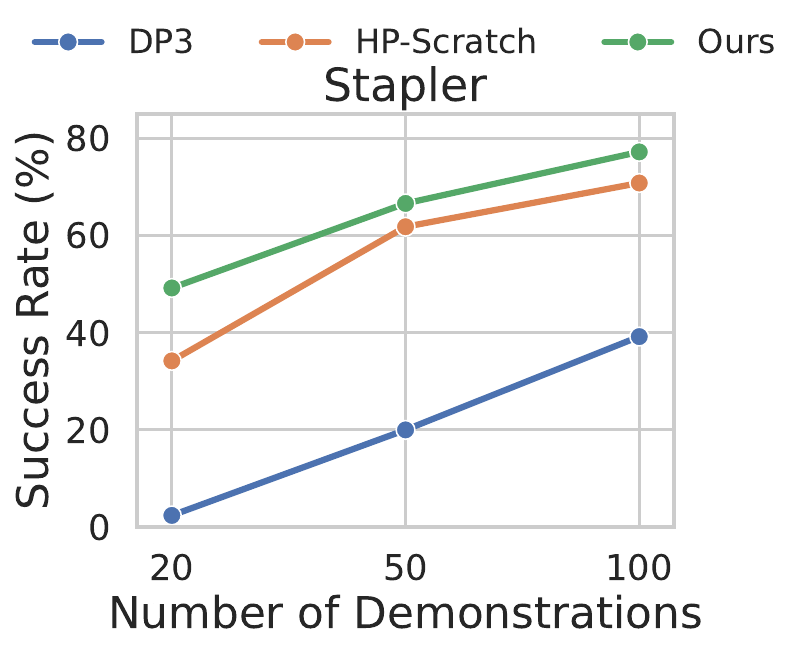}

\captionof{figure}{\small{Sample efficiency of our method compared with baselines.}}
\label{fig:stapler-scaling}
\end{minipage}

\end{figure*}

\textbf{Q1: Does hierarchical policy representation benefit performance?} We study the effect of the hierarchical policy representation. The results are shown in Table~\ref{tab:main_results}. The first three rows are end-to-end methods, while the bottom rows use hierarchy in the pipeline. We also compare to the Dex1B baseline in Table~\ref{tab:more-ablations}. Our results show that hierarchical policies provide consistent and significant performance gains over all evaluated end-to-end methods. In Figure~\ref{fig:stapler-scaling}, the results across 20, 50, and 100 demonstrations indicate that hierarchical policies are more sample-efficient than the end-to-end DP3 baseline. We observe that end-to-end policies often localize the object imprecisely, while hierarchical policies achieve more precise object localization and grasp alignment. For example, in the spray bottle task, the hand must align with the bottle orientation to establish a stable grasp. However, DP3 tends to grasp the object with a noticeable offset, which often leads to grasp failure. 



\textbf{Q2: What is the effect of low-level policy pretraining in the system?} We study the effect of low-level policy pretraining with \texttt{G2D-Pretrain}, shown in Table~\ref{tab:main_results}. Compared with \textit{HP-Scratch} which does not incorporate grasp pretraining, our pretrained model achieves higher average success rates for all demonstration budgets. We find that the benefit of pretraining varies across tasks, which can be attributed to variance in task difficulty, demonstration coverage, and the learned low-level skills of the pretrained policy. 
We also observe that pretraining provides moderate gains for end-to-end policies, as shown by the comparison between \textit{DP3} and \textit{DP3-PT}, which shows pretraining can provide a useful initialization. We additionally find that our policy performs better when conditioned on 16 keypoints rather than coarse fingertip keypoints, showing that the four-point sub-goal representation in prior work~\cite{wang2025articubot,krishna2026ghost} is not sufficient in our setting.
See Appendix D.1 for details.

\textbf{Q3: How does our method compare with VLA methods?} UniDex-VLA~\cite{zhang2026unidex} is a recent VLA specifically designed
for dexterous manipulation, which is more closely matched to our setting than
general-purpose VLAs. As shown in Table~\ref{tab:object-stage-performance}, UniDex-VLA achieves high contact success
and can often establish a grasp, but its performance drops substantially in
subsequent stages that require lifting, precise alignment, and articulation. In contrast, the hierarchy in our method enables the
low-level controller to directly leverage large-scale grasp data and provides
an explicit geometric representation for transitions between grasping,
alignment, and tool actuation.
Our goal is therefore complementary to that of general-purpose VLAs. VLAs enable broad semantic and task generalization, whereas we
focus on data-efficient learning of contact-rich dexterous manipulation from
limited downstream demonstrations. When precise multi-finger coordination and
functional grasp transitions are the primary bottleneck, our results suggest
that explicitly structuring the policy and leveraging dexterous grasp priors
can be beneficial relative to directly fine-tuning a VLA.

\textbf{Q4: Which components of the synthetic pretraining trajectories are essential?}
We further study which aspects of \texttt{G2D-Pretrain} are important by varying both
the amount and composition of the synthesized trajectories, shown in Table~\ref{tab:more-ablations}. Reducing the pretraining set from 355k to 215k trajectories (\textit{PT-215k}) and removing individual stages from the synthesized trajectories (\textit{PT w/o Appr.} and \textit{PT w/o Lift}) both lead to worse performance. These results support our design choice of converting static grasp annotations
into complete trajectories: the approach and post-grasp stages provide useful
supervision for learning how to reach, establish, and maintain hand--object
interactions during downstream manipulation.

\textbf{Q5: How does G2D-Pretrain compare to other large-scale grasp datasets for pretraining?} We compare our method against pretraining the low-level policy with an existing grasp demonstration dataset, GraspXL~\cite{zhang2024graspxl}. As is shown in Table~\ref{tab:main_results}, pretraining with \texttt{G2D-Pretrain} consistently improves performance across tasks compared to pretraining with GraspXL. 
Although GraspXL provides large-scale grasping motions across many objects, its diversity is primarily induced through sampled motion objectives such as graspable regions and approach directions. As a result, pretraining on GraspXL may bias the policy toward a narrower set of object-interaction patterns, but do not cover the functional contact modes required by downstream articulated tool use. In contrast, Dexonomy explicitly covers diverse taxonomy-level grasp types, yielding broader variation in hand configurations and contact patterns. This grasp-type diversity may better match our downstream tasks, where success often depends on selecting functionally appropriate grasps rather than only achieving stable object acquisition.









\subsection{Real World Results}

\begin{table*}[t]
\centering
\scriptsize
\setlength{\tabcolsep}{5pt}
\begin{tabular}{
lccc|ccc|ccc||
>{\columncolor{gray!10}}c
>{\columncolor{gray!10}}c
>{\columncolor{gray!10}}c
}
\toprule
& \multicolumn{3}{c|}{Syringe}
& \multicolumn{3}{c|}{Spray Bottle}
& \multicolumn{3}{c||}{Scissors}
& \multicolumn{3}{>{\columncolor{gray!10}}c}{\textbf{Avg.}} \\
\cmidrule(lr){2-4}
\cmidrule(lr){5-7}
\cmidrule(lr){8-10}
\cmidrule(l){11-13}
Task Stage
& Grasp & Align & Trigger
& Grasp & Align & Trigger
& Grasp & Align & Trigger
& Grasp & Align & Trigger \\
\midrule
DP3
& 44.4 & 5.6 & 5.6
& 27.8 & 16.7 & 5.6
& 44.4 & 38.9 & 11.1
& 38.9 & 20.4 & 7.4 \\

HP-Scratch
& 55.6 & 44.4 & 33.3
& 61.1 & 44.4 & 22.2
& \textbf{66.7} & 38.9 & 22.2
& 61.1 & 42.6 & 25.9 \\

Ours
& \textbf{88.9} & \textbf{72.2} & \textbf{38.9}
& \textbf{88.9} & \textbf{83.3} & \textbf{38.9}
& 55.6 & \textbf{44.4} & \textbf{44.4}
& \textbf{77.8} & \textbf{66.6} & \textbf{40.7} \\
\bottomrule
\end{tabular}
\caption{Success rate (\%) across three real world tasks with 50 demonstrations. The Avg. columns report the average over all tasks. We use 18 evaluation episodes in each setting.}
\label{tab:real_world_results}
\end{table*}

\begin{figure}[t]
    \centering
    \includegraphics[width=\linewidth]{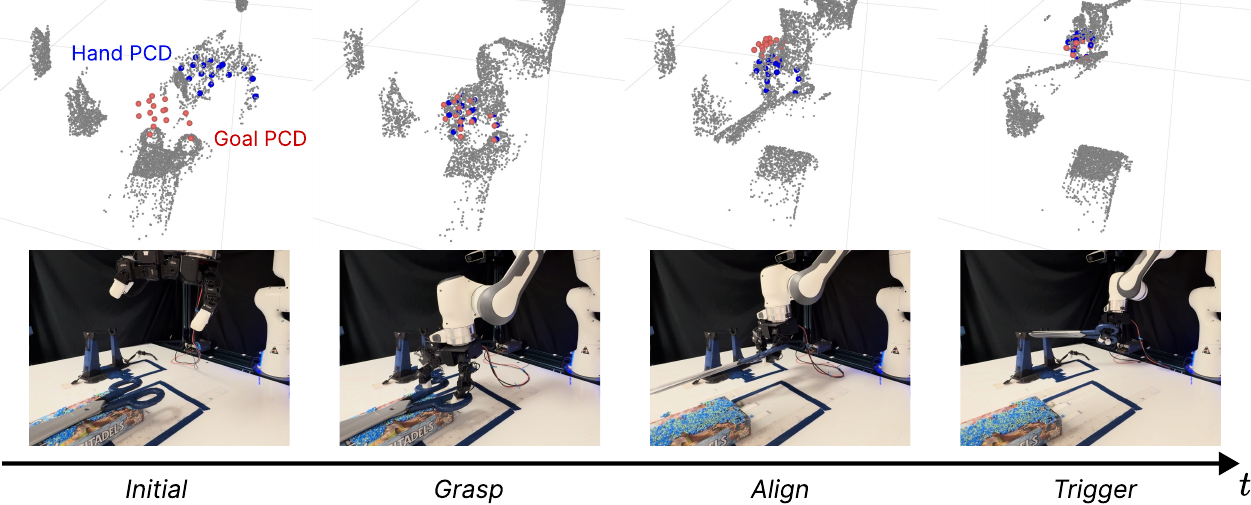}
    \caption{Visualization of high-level policy’s sub-goal predictions during real-world deployment.}
    \label{fig:high-level-vis}
\end{figure}

We further evaluate our method on real-world articulated tool use tasks, shown in Table~\ref{tab:real_world_results}. 
Our method consistently outperforms both the end-to-end DP3 baseline and the hierarchical policy trained from scratch across most task stages. In particular, our method achieves the highest average triggering success rate, improving full-task success by 33.3 percentage points over DP3. These results suggest that grasp pretraining improves not only initial grasp acquisition, but also downstream alignment and articulated tool actuation in real-world deployment. We also visualize our high-level policy's sub-goal predictions during deployment in Figure~\ref{fig:high-level-vis}.










\section{Conclusion}
\label{sec:conclusion}

In this paper, we show that large-scale grasp synthesis datasets can support dexterous downstream manipulation beyond conventional grasping. We introduce \texttt{DexCraft} for articulated tool use evaluation and \texttt{G2D-Pretrain} for grasp prior pretraining. 
By integrating these priors into a hierarchical policy framework, our method improves learning of contact-rich tool use behaviors. Our results suggest that large-scale grasp datasets can serve not only as resources for grasp generation, but also as foundations for learning more general dexterous manipulation skills.
We hope that \texttt{DexCraft} and \texttt{G2D-Pretrain} will support future research on data-driven dexterous manipulation.


\section{Limitations and Future Work}



Our work has several limitations that suggest directions for future research. 
Our hierarchical framework relies on manually labeled sub-goals, which may limit scalability. Future work could explore automatic sub-goal discovery.
Each \texttt{DexCraft} task currently uses a single object instance, thus expanding object diversity would better evaluate generalization.
In addition, our simulation demonstrations are collected via keyboard teleoperation, and higher-quality data from improved teleoperation or RL refinement may further improve performance.
Finally, larger-scale real-world evaluation with tactile feedback and online adaptation could improve robustness.


\clearpage
\acknowledgments{This material is based upon work supported by the Toyota Research Institute. This material is based upon work supported by ONR MURI N00014-24-1-2748. This material is based upon work supported by the National Science Foundation under NSF CAREER Grant No. IIS-2046491. This material is based on work that is partially funded by an unrestricted gift from Google. David Held holds concurrent appointments at CMU and as an Amazon Scholar. This paper describes work performed at CMU and is not associated with Amazon. We thank Kenny Shaw for hardware support, Amber Li and Kallol Saha for their valuable feedback on the manuscript.}


\bibliography{example}  

\clearpage
\newpage
\appendix

\addcontentsline{toc}{section}{Appendix} 
\part{Appendix} 
\parttoc 

\section{Benchmark Details}
We propose the \texttt{DexCraft} benchmark that contains articulated tool use tasks. In this section, we provide more details on our task assumption and descriptions of task setups.

\subsection{Details on Decomposing Tasks into Sub-tasks}
We first detail our assumption mentioned in Section~\ref{section:assumption}. We assume that the dexterous manipulation tasks in question consist of a sequence of sub-tasks. Following MimicGen~\cite{mandlekar2023mimicgen}, we assume that the sub-tasks are object-centric, i.e. the manipulation in each sub-task is relative to a single object's coordinate frame.

Prior hierarchical policy learning methods~\cite{wang2025articubot,krishna2026ghost} decompose tasks using either gripper open-close heuristics or manual annotation, while MimicGen~\cite{mandlekar2023mimicgen} splits tasks into object-centric sub-tasks where the robot motion is relative to an object. Unlike prior parallel-jaw gripper settings, where the end effector is often abstracted by a binary open/close state, our dexterous hand can realize diverse grasp types and multi-finger configurations. We therefore adopt a hybrid object-centric and grasp-centric decomposition, where each sub-task captures a transition in either the object pose or the robot’s grasp configuration. For example, moving the articulated tool to its target pose is object-centric, whereas grasping and triggering the tool are grasp-centric because they require different finger configurations and contact patterns.

We assume access to metrics that allow the end of each stage to be detected automatically. In this case, we can easily detect each stage, decompose data trajectories, and post-process the demonstration dataset. More specifically, we decompose each articulated tool-use task in the benchmark into three stages: grasping, aligning, and triggering.
We define stage success as follows: 1) \textit{Grasping}: At least two fingers are in contact with the object, with contact force greater than 0.5 $\mathrm{N}$; 2) \textit{Aligning}: the object is within 0.04 $\mathrm{m}$ of the target position and within 30 degrees of the target orientation; 3) \textit{Triggering}: the articulated joint reaches a task-specific threshold. 

\subsection{Task Details}
We visualize the tasks in Figure~\ref{fig:benchmark}; videos are available on our project website: \href{\web}{\web}. As is mentioned in Section~\ref{section:benchmark}, we use delta end-effector control for the arm and delta joint position control for the hand. The control frequency is 20 Hz. We implement two RGB-D cameras in the simulation environment.

\textbf{Spray bottle.} Grasp a spray bottle, move it in front of a cloth, and squeeze the trigger. The spray bottle is initialized within a $0.1\,\mathrm{m} \times 0.1\,\mathrm{m}$ region on the table with a random rotation about the $z$-axis sampled from $(-30^\circ, 30^\circ)$. The goal is randomly chosen within a $0.1\,\mathrm{m} \times 0.1\,\mathrm{m}$ region with a random rotation about the $z$-axis sampled from $(-30^\circ, 30^\circ)$ and a random rotation about the $y$-axis sampled from $(-60^\circ, 0^\circ)$. The height of the goal is sampled $0.1\,\mathrm{m}$ to $0.3\,\mathrm{m}$ above the current object $z$ position. 

\textbf{Lighter.} Grasp a flip-top lighter, move it in front of a reference object, and flip open the lid. The lighter is initialized within a $0.1\,\mathrm{m} \times 0.1\,\mathrm{m}$ region on the table with a random rotation about the $z$-axis sampled from $(-30^\circ, 30^\circ)$. The goal is randomly chosen within a $0.1\,\mathrm{m} \times 0.1\,\mathrm{m}$ region with a random rotation about the $z$-axis sampled from $(-30^\circ, 30^\circ)$ and a random rotation about the $y$-axis sampled from $(-15^\circ, 15^\circ)$. The height of the goal is sampled $0.1\,\mathrm{m}$ to $0.3\,\mathrm{m}$ above the current object $z$ position. 

\textbf{Dispenser.} Grasp a pump dispenser, move it to a reference object, and press the pump. The dispenser is initialized within a $0.1\,\mathrm{m} \times 0.1\,\mathrm{m}$ region on the table with a random rotation about the $z$-axis sampled from $(-30^\circ, 30^\circ)$. The goal is randomly chosen within a $0.1\,\mathrm{m} \times 0.1\,\mathrm{m}$ region with a random rotation about the $z$-axis sampled from $(-30^\circ, 30^\circ)$ and a random rotation about the $y$-axis sampled from $(-60^\circ, 0^\circ)$. The height of the goal is sampled $0.1\,\mathrm{m}$ to $0.3\,\mathrm{m}$ above the current object $z$ position. 

\textbf{Pliers.} Grasp a pair of pliers, move them to a reference object positioned between the jaws, and close the jaws. The pliers are placed on a $0.04\,\mathrm{m}\times 0.04\,\mathrm{m}\times 0.03\,\mathrm{m}$ box and initialized within a $0.1\,\mathrm{m} \times 0.1\,\mathrm{m}$ region with a random rotation about the $z$-axis sampled from $(-30^\circ, 30^\circ)$. The goal is randomly chosen within a $0.1\,\mathrm{m} \times 0.1\,\mathrm{m}$ region with a random rotation about the $z$-axis sampled from $(-30^\circ, 30^\circ)$ and a random rotation about the $y$-axis sampled from $(-60^\circ, 0^\circ)$. The height of the goal is sampled $0.1\,\mathrm{m}$ to $0.3\,\mathrm{m}$ above the current object $z$ position.

\textbf{Stapler.} Grasp a stapler, move it to a reference object between the top arm and the base, and press the top arm. The stapler is initially placed lying on the table for easier grasping, with its position sampled within a $0.1\,\mathrm{m} \times 0.1\,\mathrm{m}$ region and its rotation about the $z$-axis sampled from $(-30^\circ, 30^\circ)$. The goal is randomly chosen within a $0.1\,\mathrm{m} \times 0.1\,\mathrm{m}$ region with a random rotation about the $z$-axis sampled from $(-30^\circ, 30^\circ)$ and a random rotation about the $y$-axis sampled from $(-15^\circ, 15^\circ)$. The height of the goal is sampled $0.1\,\mathrm{m}$ to $0.3\,\mathrm{m}$ above the current object $z$ position. 

\textbf{Pen.} Grasp a pen, move it to a reference object near the pen tip, and click the button. The pen is placed vertically in a $0.06\,\mathrm{m}\times 0.06\,\mathrm{m}\times 0.05\,\mathrm{m}$ stand and initialized within a $0.1\,\mathrm{m} \times 0.1\,\mathrm{m}$ region on the table with a random rotation about the $z$-axis sampled from $(-30^\circ, 30^\circ)$. The goal is randomly chosen within a $0.1\,\mathrm{m} \times 0.1\,\mathrm{m}$ region with a random rotation about the $z$-axis sampled from $(-30^\circ, 30^\circ)$ and a random rotation about the $y$-axis sampled from $(-60^\circ, 0^\circ)$. The height of the goal is sampled $0.1\,\mathrm{m}$ to $0.3\,\mathrm{m}$ above the current object $z$ position. 

\subsection{Data Generation Details}
We collect 5 demos through manual keyboard teleoperation for each task. We then use MimicGen~\cite{mandlekar2023mimicgen} to generate more demos. 
For the \textit{grasping} sub-task, the reference object is the tool to be manipulated and we set the sub-task offset range to be 5 to 10 steps. 
For the \textit{aligning} sub-task, the reference object is the goal instance and we set the sub-task offset range to be 5 to 10 steps. 
For the \textit{triggering} sub-task, the reference object is also the goal instance.
We use 50 interpolation steps between sub-tasks. The data generation process has approximately 10\% success rate on average across tasks.

\section{Method Details}
\subsection{Implementation Details of High-level Policy}
We use PointNet++~\cite{qi2017pointnetplusplus} as the model backbone of the high-level policy. We append one-hot vectors as labels to each point to distinguish the camera point cloud and the current hand keypoints, and concatenate them together as input.
We model the high-level sub-goal distribution as a dense Gaussian mixture over scene points. For each point $p_i$ in the input point cloud, the network predicts displacements $\Delta_i\in\mathbb{R}^{16\times3}$ to the 16 hand keypoints and a mixture logit $w_i$. Each point defines a candidate sub-goal $\mu_i=p_i+\Delta_i$, and the logits are normalized into mixture weights $\pi_i$. Given the ground-truth sub-goal $G$, we minimize the negative log-likelihood
\[
\mathcal{L}_{\mathrm{GMM}}(G)
=
-\log
\sum_i
\pi_i
\mathcal{N}
\left(
\mathrm{vec}(G);
\mathrm{vec}(p_i+\Delta_i),
\sigma^2 I
\right).
\]
We sum this loss over multiple fixed variances $\sigma\in\Sigma$ and add a uniform-mixture auxiliary loss, which trains point-wise displacement predictions even for points assigned low mixture weights:
\[
\mathcal{L}_{\mathrm{high}}
=
\sum_{\sigma\in\Sigma}
\left(
\mathcal{L}_{\mathrm{GMM}}^\sigma
+
\lambda \mathcal{L}_{\mathrm{uniform}}^\sigma
\right).
\]

At inference time, the high-level policy samples a sub-goal from the predicted dense GMM. 
For each scene point $p_i$, the network predicts a displacement $\Delta_i \in \mathbb{R}^{16 \times 3}$ and a mixture logit $w_i$. 
After normalizing the logits into mixture weights $\pi_i$ with a softmax, we sample an anchor point $i \sim \mathrm{Categorical}(\pi)$ and use the corresponding mean
\[
\hat{G} = p_i + \Delta_i
\]
as the predicted 16-keypoint hand sub-goal. 
We use the selected Gaussian mean directly, without sampling additional Gaussian noise. Full hyperparameters are listed in Table~\ref{tab:high-level-param}.

\begin{table}[t]
\caption{High-level policy hyperparameters.}
\label{tab:high-level-param}
\centering
\begin{tabular}{ll}
\toprule
\textbf{Hyperparameter} & \textbf{Value} \\
\midrule
\multicolumn{2}{l}{\textbf{PointNet++}} \\
PointNet++ type & MSG \\
Number of set abstraction layers & 6 \\
Number of feature propagation layers & 6 \\
Set abstraction points & 1024, 512, 256, 128, 64, 16 \\
MSG radii & $[0.025,0.05] \rightarrow [0.8,1.6]$ across 6 SA layers \\
MSG neighbors & [16, 32] \\
Set abstraction MLP dims & 32/64, 128/128, 256/256, 512/512, 512/512, 512/512 \\
Feature propagation MLP dims & 512, 512, 256, 256, 128, 128 \\
Output head & Conv1D(128, 128) + BN + ReLU + Conv1D(128, $C$) \\
\midrule
\multicolumn{2}{l}{\textbf{Dense Per-Patch GMM}} \\
Fixed variances & \texttt{[0.01, 0.05, 0.1, 0.25, 0.5]} \\
Uniform weights coefficient & 0.1 \\
\midrule
\multicolumn{2}{l}{\textbf{Training}} \\
Epochs & 50 \\
Learning Rate & $3 \times 10^{-4}$ \\
Optimizer & AdamW\\
Batch Size & 32 \\
\bottomrule
\end{tabular}
\end{table}

\subsection{Implementation Details of Low-level Policy}
\label{app:low-level-detail}
Following prior work~\cite{wang2025articubot}, we use a modified DP3 encoder for the low-level goal-conditioned policy. The encoder takes as input the scene point cloud, the current hand point cloud, the predicted goal hand point cloud from the high-level policy, and the robot proprioceptive state. To distinguish different point types, we append one-hot labels to the input points, indicating whether a point belongs to the scene, the current hand, or the goal hand configuration. The point cloud input is then encoded with an Act3D~\cite{gervet2023act3d}-style cross-attention module. Specifically, scene points are first embedded with a point-wise MLP and 3D rotary positional encoding. Current hand keypoints are used as queries to attend to scene features, allowing the policy to extract task-relevant object context around the hand. A second cross-attention module relates the current hand keypoints to the goal hand keypoints, providing an explicit representation of the desired hand motion. The resulting point-cloud feature is concatenated with a proprioceptive state embedding and used as the observation feature for the DP3 diffusion policy. This modification allows the low-level policy to condition action generation on both the current dexterous hand configuration and the predicted 16-keypoint sub-goal, rather than relying only on a global scene representation. Hyperparameters are listed in Table~\ref{tab:low_level_hparams}.

\begin{table}[t]
\centering
\small
\caption{Low-level policy hyperparameters.}
\label{tab:low_level_hparams}
\begin{tabular}{ll}
\toprule
\textbf{Hyperparameter} & \textbf{Value} \\
\midrule
\multicolumn{2}{l}{\textbf{DP3}} \\
Conditioning type & FiLM \\
Observation steps & 2 \\
Prediction horizon & 16 \\
Action steps & 4 \\
Diffusion steps during training & 100 \\
Diffusion steps during inference & 10 \\
Noise schedule & DDIM, squared cosine \\
Diffusion step embedding dim & 128 \\
U-Net down dimensions & [512, 1024, 2048] \\
Kernel size & 5 \\
Number of groups & 8 \\
\midrule
\multicolumn{2}{l}{\textbf{Point Cloud Encoder}} \\
Scene points & 1024 \\
End-effector keypoints & 16 \\
Point feature dim & 6 \\
Point cloud color & False \\
Encoder output dim & 64 \\
State MLP dims & [64, 64] \\
\midrule
\multicolumn{2}{l}{\textbf{Training}} \\
Optimizer & AdamW \\
Learning rate & $1\times 10^{-4}$ \\
AdamW betas & [0.95, 0.999] \\
Weight decay & $1\times 10^{-6}$ \\
Batch size & 128 \\
Training epochs & 500 \\
Learning rate schedule & cosine \\
Warmup steps & 500 \\
EMA max decay & 0.9999 \\
\bottomrule
\end{tabular}
\end{table}

\subsection{Implementation Details of Generating \texttt{G2D-Pretrain}}
To convert static grasp data into policy pretraining trajectories, we generate demonstration rollouts from Dexonomy grasp annotations~\cite{chen2025dexonomy} in the ManiSkill3~\cite{taomaniskill3} simulator. Each Dexonomy sample provides a sequence of hand-object configurations corresponding to pre-grasp, grasp, and squeeze states. We first convert these configurations into robot action key-frames by extracting the wrist pose and Shadow-hand joint positions and applying a random spatial offset to both the object pose and target hand poses. This produces randomized grasping scenes while preserving the relative hand-object grasp geometry.

For each grasp instance, we construct a key-frame trajectory consisting of a randomized initial hand pose, an open-hand pose aligned with the target pre-grasp pose, the pre-grasp configuration, the grasp configuration, the squeeze configuration, and a final lifted pose. We execute these key-frames in a simulated environment with point-cloud observations and joint-position delta control. For the approach phase, we use a motion planner with the observed object point cloud for collision checking; for later phases, we linearly interpolate between key-frames and smoothly interpolate the finger joints. The resulting relative joint-position commands are executed in the environment to collect observations, actions, and robot states.

\begin{figure}
    \centering
    \includegraphics[width=0.8\linewidth]{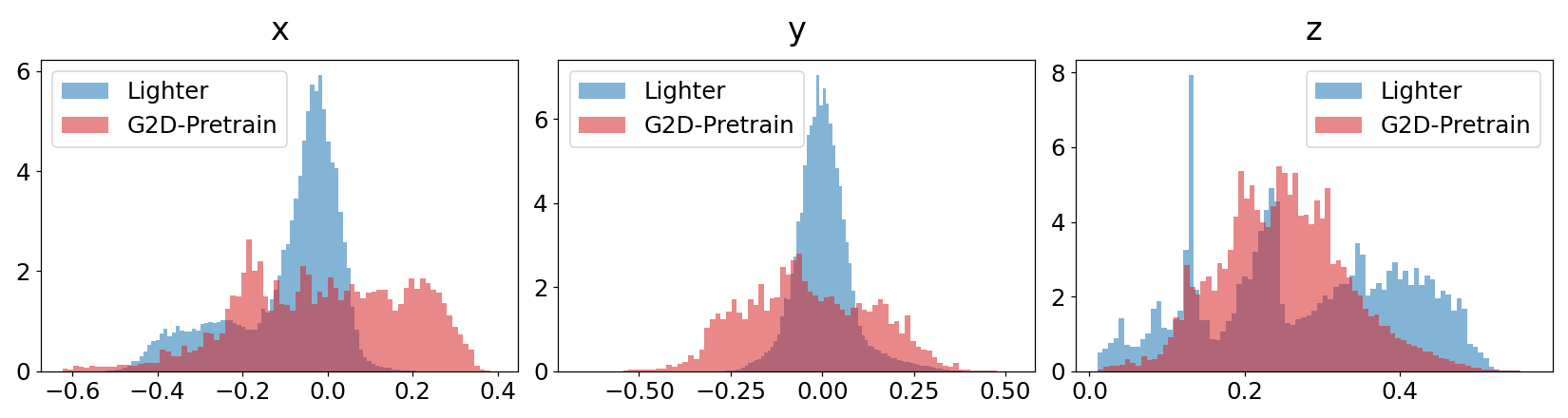}
    \caption{Point cloud statistics of \texttt{G2D-Pretrain} compared with a downstream task.}
    \label{fig:viz-pcd-hist-dex}
    \vspace{-1em}
\end{figure}

\begin{figure}
    \centering
    \includegraphics[width=0.8\linewidth]{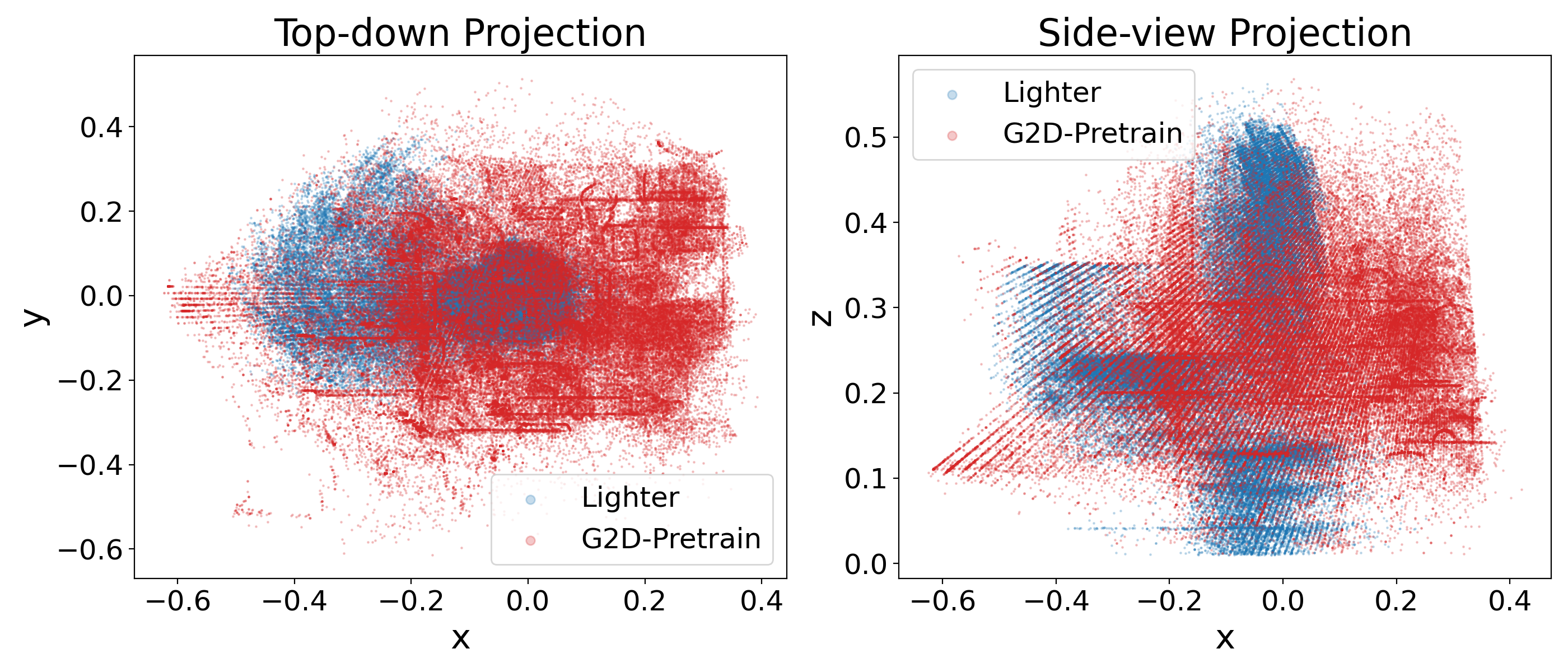}
    \caption{Workspace coverage comparison between \texttt{G2D-Pretrain} and a downstream task, visualized as point-cloud projections onto the $x$-$y$ and $x$-$z$ planes.}
    \label{fig:viz-pcd-proj-dex}
\end{figure}

We filter generated trajectories based on whether the simulator reports a valid grasp. For each valid trajectory, we additionally construct stage-wise goal hand point clouds from the hand point cloud at key transition indices, so that the low-level policy can be trained with goal-conditioned observations. We save each trajectory as arrays of actions, proprioceptive states, scene point clouds, current hand point clouds, and goal hand point clouds. This process turns taxonomy-level grasp annotations into closed-loop-compatible grasp demonstrations for low-level policy pretraining. Table~\ref{tab:grasp_type_counts} summarizes the distribution of generated pretraining trajectories across grasp types. We further compare point-cloud statistics and workspace coverage between \texttt{G2D-Pretrain} and a downstream \texttt{DexCraft} task in Figures~\ref{fig:viz-pcd-hist-dex} and~\ref{fig:viz-pcd-proj-dex}. 
These visualizations show that \texttt{G2D-Pretrain} spans a broad workspace and covers diverse hand-object configurations, providing useful pretraining coverage for downstream dexterous manipulation.

\begin{table}[t]
\centering
\small
\setlength{\tabcolsep}{6pt}
\caption{Number of generated trajectories for each grasp type in \texttt{G2D-Pretrain}.}
\label{tab:grasp_type_counts}
\begin{tabular}{lc|lc}
\toprule
\textbf{Grasp type} & \textbf{\# Traj.} & \textbf{Grasp type} & \textbf{\# Traj.} \\
\midrule
1\_Large\_Diameter  & 37,003 & 17\_Index\_Finger\_Extension & 12,159 \\
2\_Small\_Diameter  & 15,176 & 18\_Extensior\_Type & 11,590 \\
3\_Medium\_Wrap  & 16,453 & 20\_Writing\_Tripod & 5,669 \\
4\_Adducted\_Thumb  & 4,293 & 22\_Parallel\_Extension & 11,458 \\
5\_Light\_Tool  & 3,783 & 23\_Adduction\_Grip & 2,342 \\
6\_Prismatic\_4\_Finger  & 2,970 & 24\_Tip\_Pinch & 6,382 \\
7\_Prismatic\_3\_Finger  & 3,323 & 25\_Lateral\_Tripod & 6,801 \\
8\_Prismatic\_2\_Finger  & 2,585 & 26\_Sphere\_4\_Finger & 17,516 \\
9\_Palmar\_Pinch  & 1,428 & 27\_Quadpod & 423 \\
10\_Power\_Disk & 459 & 28\_Sphere\_3\_Finger & 24,354 \\
11\_Power\_Sphere & 23,707 & 29\_Stick & 21,628 \\
12\_Precision\_Disk & 22,407 & 30\_Palmar & 20,887 \\
13\_Precision\_Sphere & 18,850 & 31\_Ring & 15,018 \\
14\_Tripod & 5,186 & 32\_Ventral & 17,410 \\
15\_Fixed\_Hook & 11,487 & 33\_Inferior\_Pincer & 11,419 \\
16\_Lateral & 1,511 &  &  \\
\midrule
\multicolumn{3}{r}{\textbf{Total}} & \textbf{355,677} \\
\bottomrule
\end{tabular}
\end{table}

\subsection{Implementation Details of Grasp Pretraining and Downstream Fine-tuning}

We pretrain the low-level policy on \texttt{G2D-Pretrain}, which is generated with a floating 22-DoF Shadow hand. During pretraining, the policy is trained directly in the Shadow-hand action space. This allows the DP3 policy to learn general dexterous grasping and post-grasp motion priors from large-scale grasp demonstrations.

For downstream \texttt{DexCraft} tasks, the robot hand is a LEAP hand with a different action and proprioceptive state space. To transfer the pretrained low-level policy, we use a deterministic manual adapter during downstream fine-tuning and deployment.
The pretrained low-level policy operates in a canonical Shadow-hand action space consisting of a 6-DoF relative end-effector command and 22 Shadow-hand joint commands. In contrast, the LEAP-hand downstream policy uses the same 6-DoF end-effector command but only 16 hand joint commands.

During fine-tuning, each downstream LEAP action is embedded into the canonical Shadow-hand action space before computing the diffusion loss. We copy the 6-DoF end-effector command directly, map each LEAP joint to its semantically corresponding Shadow-hand joint (Table~\ref{tab:leap_shadow_mapping}), and fill unused Shadow-hand joints with zeros. This gives a Shadow-style target action that is compatible with the pretrained action dimension. During inference, the process is reversed: the policy predicts a Shadow-style action, from which we copy the end-effector command and select the corresponding hand joint entries to form the LEAP action. The adapter contains no learnable parameters; it simply preserves the pretrained Shadow-hand action interface while allowing the policy to be fine-tuned and executed on a lower-DoF LEAP hand.

\begin{table}[t]
\centering
\small
\setlength{\tabcolsep}{6pt}
\caption{Manual joint mapping from LEAP-hand actions to the canonical Shadow-hand action space. The LEAP joint indices are zero-indexed.}
\label{tab:leap_shadow_mapping}
\begin{tabular}{ccc|ccc}
\toprule
\textbf{LEAP joint} & \textbf{Shadow slot} & \textbf{Shadow joint}
& \textbf{LEAP joint} & \textbf{Shadow slot} & \textbf{Shadow joint} \\
\midrule
0  & 1  & FFJ3 & 8  & 2  & FFJ2 \\
1  & 5  & MFJ3 & 9  & 6  & MFJ2 \\
2  & 9  & RFJ3 & 10 & 10 & RFJ2 \\
3  & 18 & THJ4 & 11 & 20 & THJ2 \\
4  & 0  & FFJ4 & 12 & 3  & FFJ1 \\
5  & 4  & MFJ4 & 13 & 7  & MFJ1 \\
6  & 8  & RFJ4 & 14 & 11 & RFJ1 \\
7  & 17 & THJ5 & 15 & 21 & THJ1 \\
\bottomrule
\end{tabular}
\vspace{-1em}
\end{table}


\section{Baseline and Ablation Details}
\subsection{Diffusion-based Method Details}
\textbf{Diffusion Policy.} We include an image-based Diffusion Policy baseline to evaluate how a standard flat visuomotor diffusion policy performs when trained directly on downstream DexHier demonstrations. The baseline uses two fixed RGB camera views, denoted as \texttt{image\_0} and \texttt{image\_1}, together with a low-dimensional robot state, and predicts low-level actions without grasp pretraining or hierarchical decomposition. We train separate policies for each task and demonstration budget, and report all results using the epoch-1000 checkpoint with 500 evaluation rollouts. The key implementation and training hyperparameters are summarized in Table~\ref{tab:dp_baseline_hparams}.

\begin{table}[t]
\centering
\small
\caption{Image-based Diffusion Policy baseline hyperparameters.}
\label{tab:dp_baseline_hparams}
\begin{tabular}{ll}
\toprule
\textbf{Hyperparameter} & \textbf{Value} \\
\midrule
\multicolumn{2}{l}{\textbf{Diffusion Policy}} \\
Conditioning type & Global observation conditioning \\
Observation steps & 2 \\
Prediction horizon & 16 \\
Action steps & 8 \\
Diffusion steps during training & 100 \\
Diffusion steps during inference & 100 \\
Noise schedule & DDPM, squared cosine \\
Diffusion step embedding dim & 128 \\
U-Net down dimensions & [512, 1024, 2048] \\
Kernel size & 5 \\
Number of groups & 8 \\
\midrule
\multicolumn{2}{l}{\textbf{Image and State Encoder}} \\
RGB observations & \texttt{image\_0}, \texttt{image\_1} \\
Image resolution & $3 \times 128 \times 128$ per view \\
Image encoder & ResNet-18 \\
Image crop & Random crop to $76 \times 76$ \\
Image normalization & ImageNet normalization \\
Normalization layers & GroupNorm \\
\midrule
\multicolumn{2}{l}{\textbf{Training and Evaluation}} \\
Optimizer & AdamW \\
Learning rate & $1\times 10^{-4}$ \\
AdamW betas & [0.95, 0.999] \\
Weight decay & $1\times 10^{-6}$ \\
Batch size & 64 \\
Training epoch & 1000 \\
Learning rate schedule & cosine \\
Warmup steps & 500 \\
EMA max decay & 0.9999 \\
\bottomrule
\end{tabular}
\vspace{-1em}
\end{table}

\textbf{3D Diffusion Policy.} We evaluate with the modified DP3 described in Section~\ref{app:low-level-detail}. We replace the entry of the goal hand keypoints into zeros before feeding into the model and keep other settings the same.

\subsection{GraspXL Dataset Conversion Details}
\label{app:graspxl_conversion}

GraspXL~\cite{zhang2024graspxl} provides large-scale grasping motion trajectories for the Allegro hand. Each trajectory contains the hand motion and object motion over time. To use GraspXL as a pretraining source for our low-level policy, we replay each trajectory to build observation and action fields that are compatible with our downstream goal-conditioned policy. Specifically, each converted trajectory contains point-cloud observations generated from the object and reconstructed hand meshes, low-dimensional states from the replayed hand and wrist motion, goal hand representations from future hand configurations, and actions defined as relative wrist and hand-joint motion between consecutive timesteps.

We sample 355k trajectories across the large, medium, and small object groups in GraspXL to construct a pretraining dataset at a comparable scale to G2D-Pretrain. The sampling strategy is designed to span diverse object identities, object sizes, and grasping directions, rather than overrepresenting a narrow subset of objects or poses. This allows us to compare the two pretraining sources under a similar data-scale setting, while reducing the possibility that differences arise simply from limited object or pose coverage.

\subsection{Transfer Protocols for Pretraining and Fine-Tuning}
\label{app:transfer_protocols}

\begin{figure}
    \centering
    \includegraphics[width=0.6\linewidth]{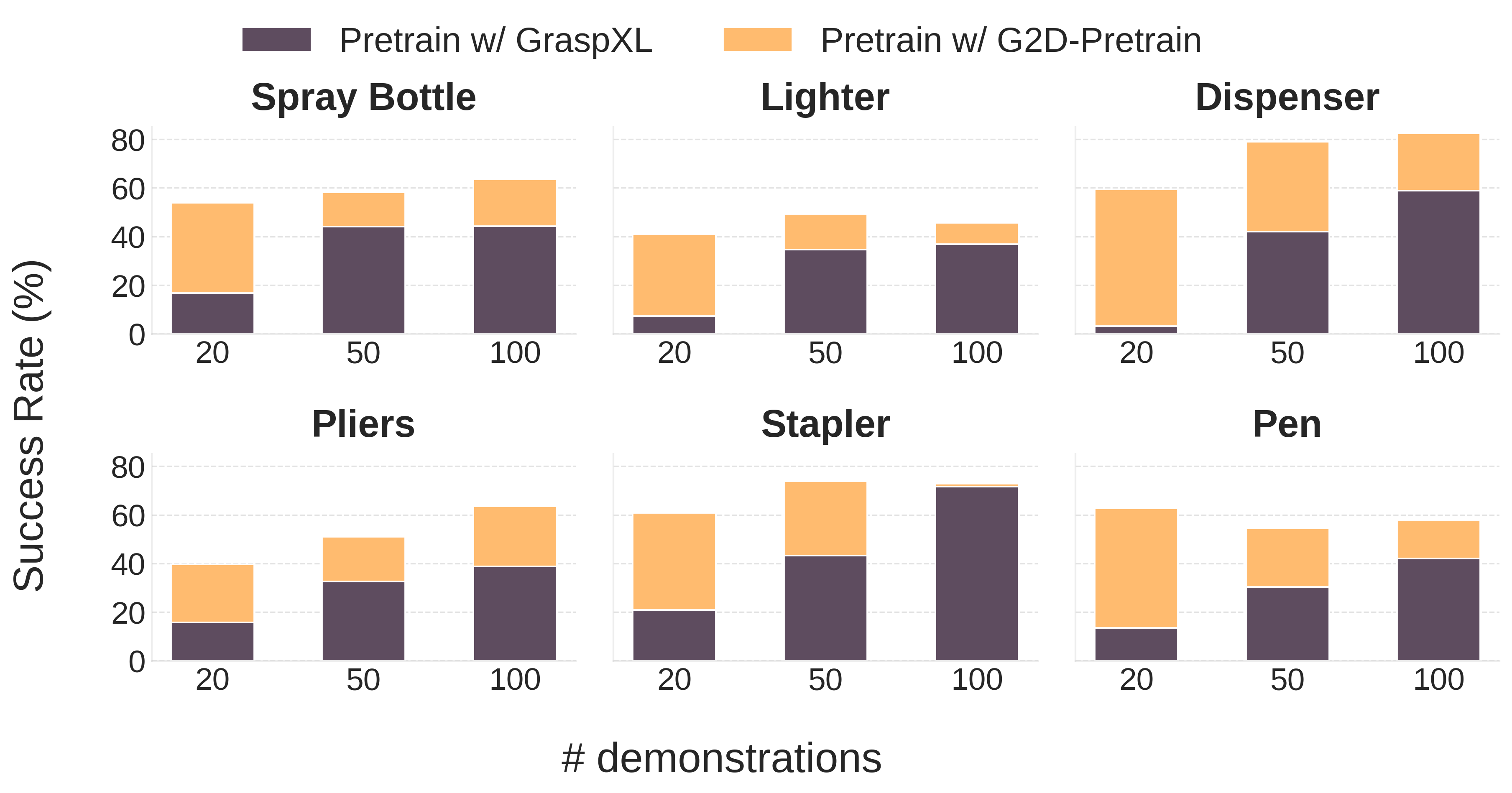}
    \caption{Performance gains from encoder-only transfer with \texttt{G2D-Pretrain} relative to GraspXL in simulation.}
    \label{fig:encoder-only transfer}
    \vspace{-0.8em}
\end{figure}

\paragraph{Encoder-only transfer for GraspXL.}
For GraspXL, we use encoder-only transfer rather than full-checkpoint transfer. Although our conversion and canonical policy interface make GraspXL compatible with the downstream low-level policy, full-checkpoint fine-tuning performs poorly in our experiments. We hypothesize that this occurs because the pretrained diffusion/action decoder remains tied to the Allegro-hand grasping distribution, while the downstream tasks differ in embodiment, state statistics, and controller dynamics. The observation encoder, however, can still provide useful visual-geometric features over point clouds and goal hand representations. We therefore transfer only the observation encoder for GraspXL and reinitialize the diffusion/action decoder during downstream fine-tuning.

\paragraph{Comparison under the same transfer protocol.}
Since GraspXL is evaluated using encoder-only transfer, we also evaluate G2D-Pretrain using the same encoder-only protocol (Figure~\ref{fig:encoder-only transfer}). This comparison isolates the effect of the pretraining source while keeping the transfer mechanism fixed. Under this matched protocol, G2D-Pretrain achieves higher downstream success than GraspXL.

We attribute this gap to the structure of the pretraining data. GraspXL primarily provides approach-and-grasp trajectories. In contrast, G2D-Pretrain covers more diverse grasp types and includes dexterous hand motion throughout the trajectory, such as grasp formation, squeeze-like finger motion, and post-grasp transport. These properties better match the interaction structure required by downstream articulated tool-use tasks, where the robot must acquire a tool, maintain stable contact, and execute functional finger motion.


\section{Analysis}

\subsection{Ablation on Keypoint Conditioning Schemes}
\label{app:ablation-keypoints}

\begin{figure}[t]
    \centering
    \vspace{-0.8em}

    \begin{minipage}[t]{0.58\linewidth}
        \centering
        \includegraphics[width=\linewidth]{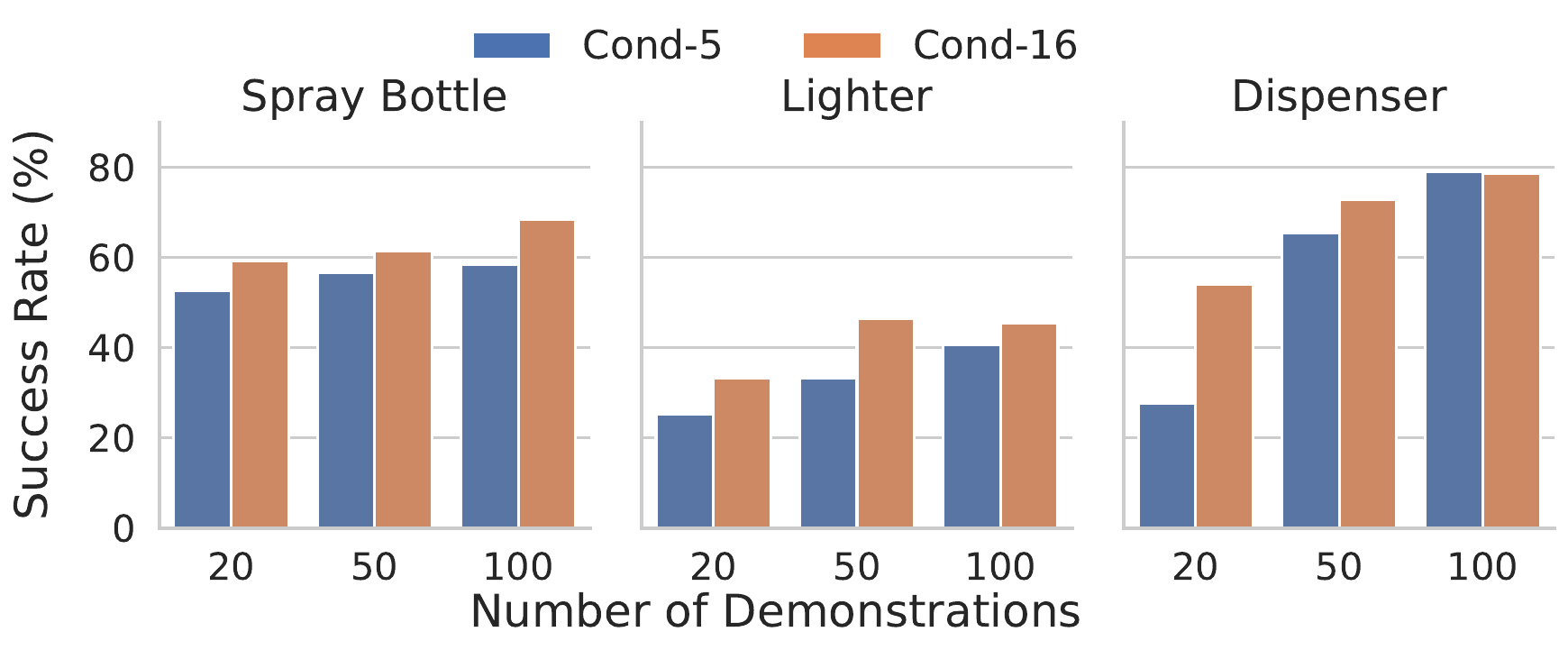}
        \captionof{figure}{Success rate (\%) of our method across three simulation tasks under different key-point conditioning schemes, comparing 5 key points with 16 key points.}
        \label{fig:conditioning-points}
    \end{minipage}
    \hfill
    \begin{minipage}[t]{0.38\linewidth}
        \centering
        \includegraphics[width=\linewidth]{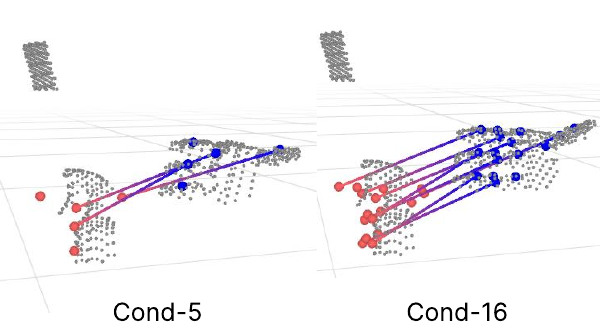}
        \captionof{figure}{Visualization of the 5-keypoint and 16-keypoint conditioning schemes.}
        \label{fig:vis-conditioning-points}
    \end{minipage}

    \vspace{-1.0em}
\end{figure}

We evaluate different forms of the 3D representation predicted by the high-level policy. Figure~\ref{fig:conditioning-points} plots the performance of our method under different key-point conditioning.
\textit{Cond-5} uses 5 key points: four points on the fingertips and one point on the palm. In contrast, \textit{Cond-16} uses 16 key points, with one point placed on each hand link (Figure~\ref{fig:vis-conditioning-points}). 
Across three simulation tasks, we find that the low-level policy benefits more from pretraining when conditioned on sub-goals that capture the full hand skeleton rather than coarse fingertip keypoints. This is likely because five key points remain insufficient to uniquely specify the complete hand configuration, leading to ambiguity for the low-level execution.

\subsection{Qualitative Analysis of Simulation Tasks}
\begin{figure}[t]
    \centering
    \includegraphics[width=0.8\linewidth]{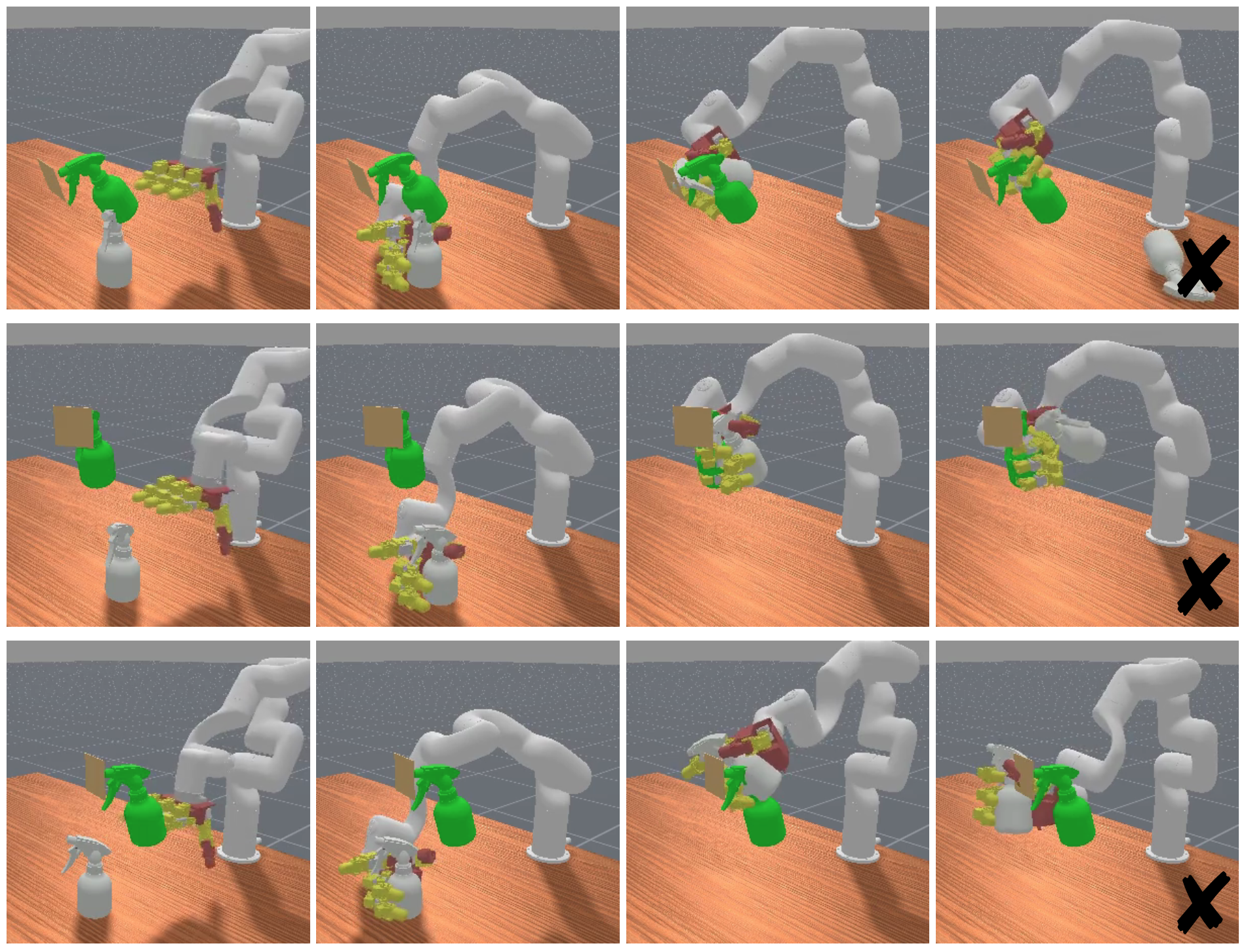}
    \caption{Example evaluation episodes of DP3 on the spray bottle task.}
    \label{app:sim-dp3}
    \vspace{-1.5em}
\end{figure}

\begin{figure}[t]
    \centering
    \includegraphics[width=0.8\linewidth]{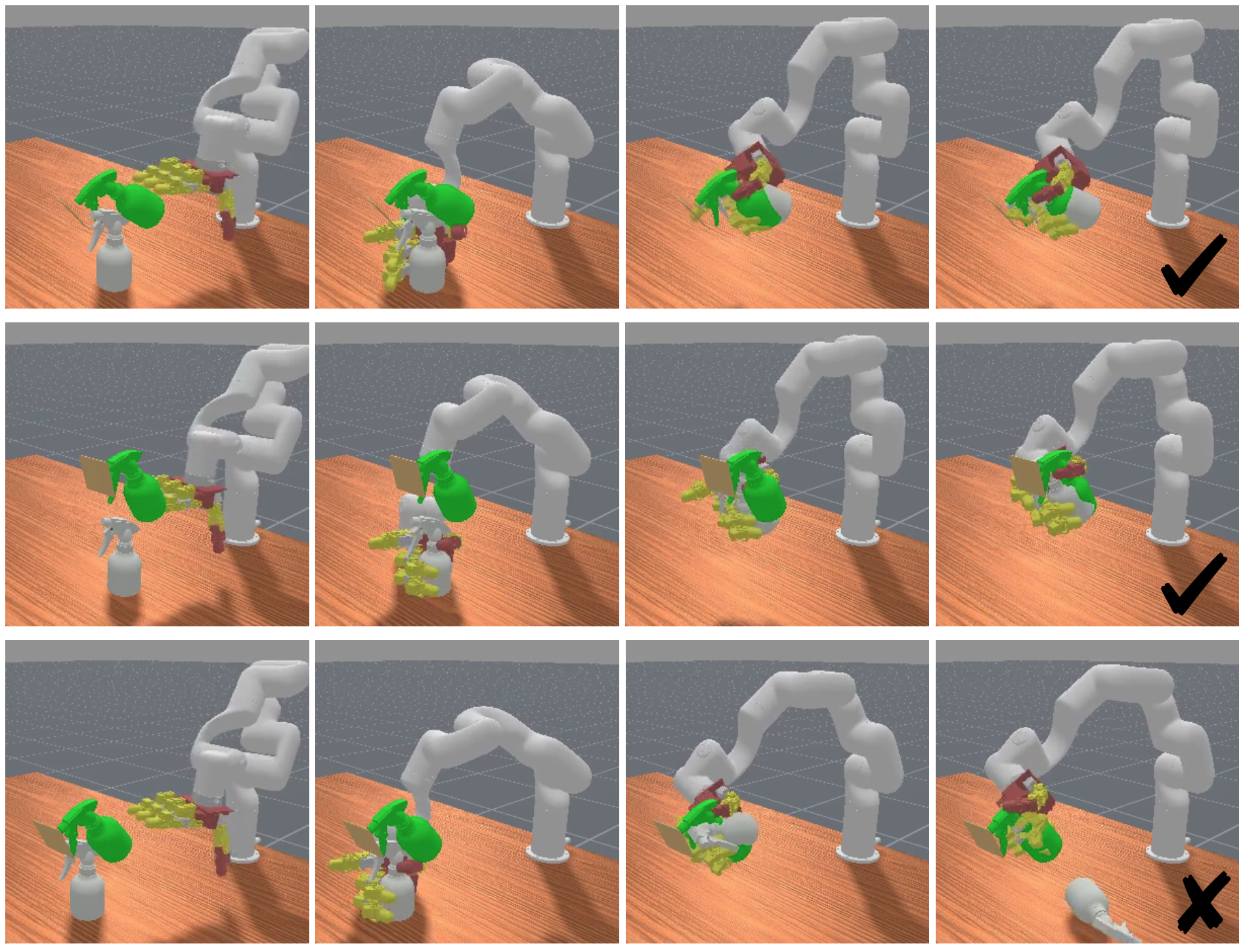}
    \caption{Example evaluation episodes of hierarchical policy learning from scratch on the spray bottle task.}
    \label{app:sim-goal}
    \vspace{-1.5em}
\end{figure}

\begin{figure}[t]
    \centering
    \includegraphics[width=0.8\linewidth]{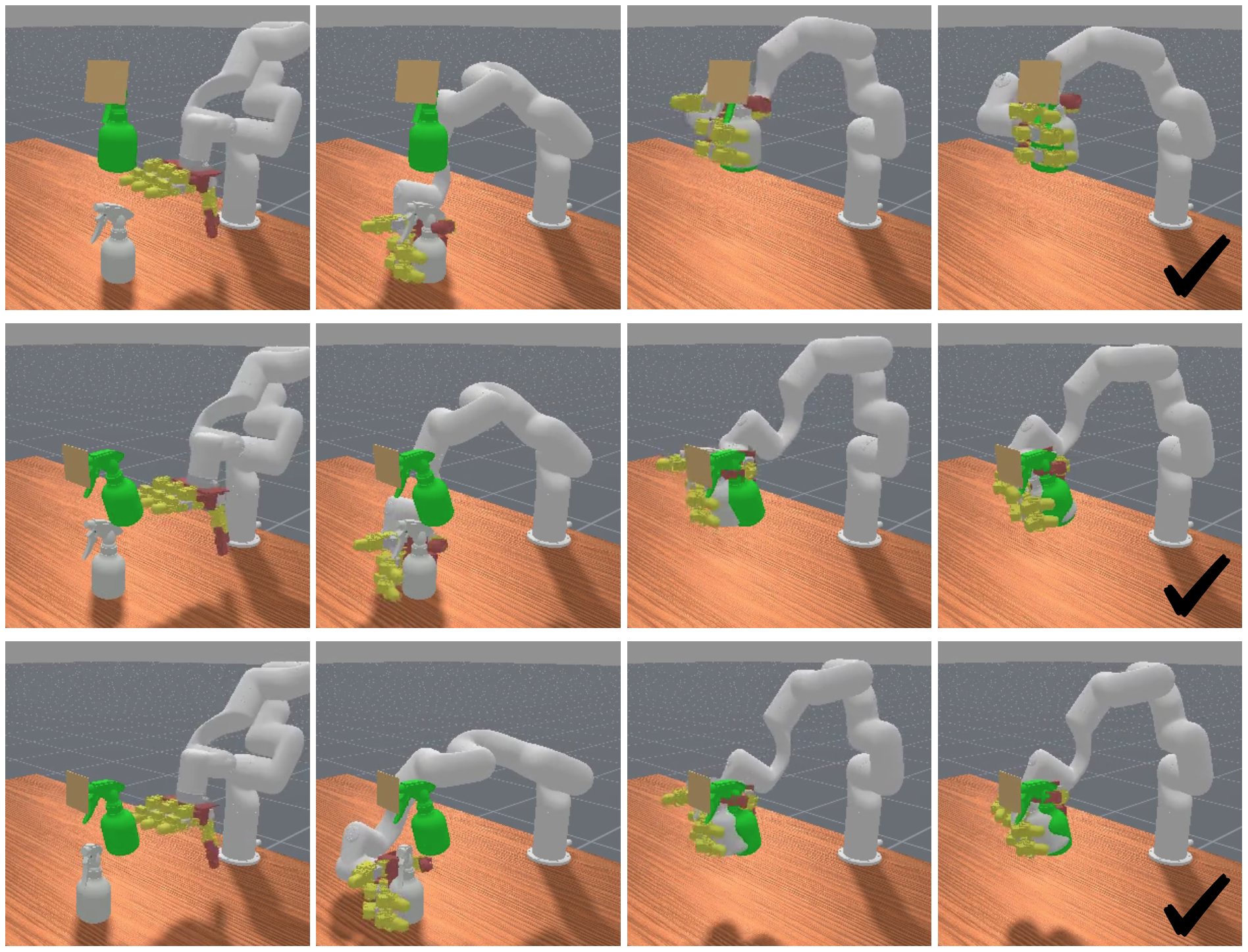}
    \caption{Example evaluation episodes of our method on the spray bottle task.}
    \label{app:sim-finetune}
    \vspace{-1.5em}
\end{figure}

Figures~\ref{app:sim-dp3},~\ref{app:sim-goal}, and~\ref{app:sim-finetune} visualize rollout episodes on the spray bottle task for DP3, hierarchical policy from scratch, and our method, respectively. 
Qualitatively, the rollouts reveal distinct failure modes across methods. The end-to-end DP3 baseline often fails at the initial grasping stage: small errors in hand-object alignment lead to unstable contact, causing the hand to slip or grasp the spray bottle in a configuration that cannot support triggering. Introducing hierarchy improves the behavior, as the high-level sub-goal provides a more structured target for the low-level controller and leads to a higher success rate than the end-to-end policy. However, when trained from scratch, the hierarchical policy still produces imperfect grasps in some cases, suggesting that task demonstrations alone are insufficient to reliably learn robust dexterous contact patterns. In contrast, our method produces more stable and task-appropriate grasps, allowing the hand to maintain contact through alignment and triggering. These observations support our quantitative results and suggest that grasp pretraining provides useful contact and hand-configuration priors for downstream articulated tool use.

\label{app:failure-case}

\subsection{High-level Policy Prediction Analysis}
\label{app:high-level-vs-oracle}

\begin{table*}[h]
\centering
\scriptsize
\setlength{\tabcolsep}{2.5pt}
\caption{Success rate (\%) across six simulation tasks with 20, 50, and 100 demonstrations. The Avg. columns report the average over all tasks. We use 500 rollout episodes in each setting.}
\begin{tabular}{lccc|ccc|ccc|ccc|ccc|ccc|>{\columncolor{gray!10}}c
>{\columncolor{gray!10}}c
>{\columncolor{gray!10}}c}
\toprule
& \multicolumn{3}{c|}{SprayBot.}
& \multicolumn{3}{c|}{Lighter}
& \multicolumn{3}{c|}{Dispenser}
& \multicolumn{3}{c|}{Pliers}
& \multicolumn{3}{c|}{Stapler}
& \multicolumn{3}{c|}{Pen}
& \multicolumn{3}{>{\columncolor{gray!10}}c}{\textbf{Avg.}} \\
\cmidrule(lr){2-4}
\cmidrule(lr){5-7}
\cmidrule(lr){8-10}
\cmidrule(lr){11-13}
\cmidrule(lr){14-16}
\cmidrule(lr){17-19}
\cmidrule(lr){20-22}
\# demo
& 20 & 50 & 100
& 20 & 50 & 100
& 20 & 50 & 100
& 20 & 50 & 100
& 20 & 50 & 100
& 20 & 50 & 100
& 20 & 50 & 100 \\
\midrule
Ours
& \textbf{59.2} & 61.4 & 68.4
& \textbf{33.2} & \textbf{46.4} & 45.4
& \textbf{54.0} & \textbf{72.8} & \textbf{78.6}
& \textbf{32.8} & \textbf{58.0} & \textbf{62.6}
& 49.2 & \textbf{66.6} & \textbf{77.2}
& 43.8 & 56.4 & 54.6
& \textbf{45.4} & \textbf{60.3} & \textbf{64.6} \\
Oracle
& 40.4 & \textbf{66.0} & \textbf{71.4}
& 26.0 & 41.0 & \textbf{48.4}
& 32.0 & 51.0 & 49.0
& 26.4 & 38.4 & 46.0
& \textbf{51.4} & 66.0 & 74.0
& \textbf{45.4} & \textbf{62.4} & \textbf{58.0}
& 36.9 & 54.1 & 57.8 \\
\bottomrule
\end{tabular}
\label{tab:main_results_2}
\vspace{-1em}
\end{table*}

We further compare our learned high-level policy with an oracle sub-goal conditioning baseline in Table~\ref{tab:main_results_2}. The oracle baseline conditions the low-level policy on ground-truth sub-goals extracted from the collected demonstrations, and switches to the next sub-goal once the end-effector reaches within a threshold of the previous one. Our method achieves performance close to oracle conditioning across most tasks and demonstration settings, and even outperforms it in several cases. This suggests that the learned high-level policy predicts effective sub-goals for downstream execution. Moreover, the oracle baseline is not necessarily an upper bound: because it follows a fixed sequence of demonstration sub-goals, it can be brittle when the object is accidentally displaced or the rollout deviates from the demonstration trajectory. In contrast, our high-level policy predicts sub-goals from the current observation, allowing it to adapt to the evolving scene state and recover from such deviations.


\end{document}